\documentclass[11pt,twocolumn,letterpaper]{article}

\usepackage{wacv}
\usepackage{times}
\usepackage{epsfig}
\usepackage{graphicx}
\usepackage{amsmath}
\usepackage{amssymb}
\usepackage{caption}

\usepackage{arydshln}
\usepackage{wrapfig}
\usepackage{subfig}
\usepackage{hyperref}
\hypersetup{
    colorlinks=true,
    linkcolor=blue,
    filecolor=magenta,
    urlcolor=cyan,
}
\DeclareMathOperator*{\argmin}{arg\,min}
\usepackage{algorithm,algpseudocode}
\usepackage{multirow}
\usepackage[normalem]{ulem}
\newcommand{\name}[0]{GPP~}

\algnewcommand\Input{\item[\textbf{Input:}]}%
\algnewcommand\Output{\item[\textbf{Output:}]}%
\algnewcommand{\To}{\textbf{To }}
\newcommand{\norm}[1]{\left\lVert#1\right\rVert}

\newcommand{\y}{\mathbf{y}}
\newcommand{\x}{\mathbf{x}}
\newcommand{\1}{\mathbf{1}}
\newcommand{\p}{\Phi}
\newcommand{\pt}{(a\Phi+b\mathbf{1})}
\newcommand{\T}{\intercal}
\usepackage[title]{appendix}

\usepackage{amsfonts}
\usepackage{amssymb}

\wacvfinalcopy
\begin{document}

\title{\textbf{Generative Patch Priors for Practical Compressive Image Recovery}}

\author{Rushil Anirudh\\
Lawrence Livermore National Laboratory,\\
Livermore, CA\\
{\tt\small anirudh1@llnl.gov}
\and
Suhas Lohit\\
Misubishi Electric Research Laboratories\\
Cambridge, MA\\
{\tt\small slohit@merl.com}
\and
Pavan Turaga\\
Arizona State University\\
Tempe, AZ.\\
{\tt\small pturaga@asu.edu}
}
\date{}
\maketitle

\begin{abstract}
   In this paper, we propose the generative patch prior (GPP) that defines a generative prior for compressive image recovery, based on patch-manifold models. Unlike learned, image-level priors that are restricted to the range space of a pre-trained generator, GPP can recover a wide variety of natural images using a pre-trained patch generator. Additionally, GPP retains the benefits of generative priors like high reconstruction quality at extremely low sensing rates, while also being much more generally applicable. We show that GPP outperforms several unsupervised and supervised techniques on three different sensing models -- linear compressive sensing with known, and unknown calibration settings, and the non-linear phase retrieval problem. Finally, we propose an alternating optimization strategy using GPP for joint calibration-and-reconstruction which performs favorably against several baselines on a real world, un-calibrated compressive sensing dataset.

\end{abstract}

\section{Introduction}
\label{sec:introduction}
Deep generative priors have proven to be an effective alternative to supervised methods in a variety of inverse problems from compressive sensing \cite{bora2017compressed}, inpainting \cite{yeh2017semantic}, to blind recovery \cite{hand2019global,asim2019blind}. These priors are expressed using Generative Adversarial Networks (GAN) \cite{GANGoodfellow} or Variational Autoencoders (VAE) \cite{kingma2013auto}, which are pre-trained to approximate the image manifold. The solution to the inverse problem is then constrained to lie on this manifold via optimizing in the latent space of the generative model. Due to their ability to characterize the image manifold better than more traditional priors like sparsity \cite{candes2008introduction} or total variation \cite{li2013efficient}, these learned priors work well in recovering the true signal, particularly under extreme sensing scenarios with very few observations. However, they are not applicable when the image to be recovered lies far away from the modes of the training distribution. As a result, generative priors today are only applicable for specific domains on which high quality GANs (and therefore better generative priors) can be trained, such as digits or faces. 

As such, a generic generative prior that is able to aid in recovering high quality, and high resolution natural images does not exist today. This is partly owing to the fact that training an unconditional GAN on a dataset like ImageNet \cite{russakovsky2015imagenet} remains a challenge due to its complexity. Even if such a generative model were trained, it is still likely to be limited in the resolution and aspect ratios of images that can be recovered. Accordingly, we hypothesize that there exists a trade-off between \emph{generalizability} of the prior (i.e., ability to recover a wide variety of images) versus its \emph{efficiency}, (i.e. effectiveness in recovering accurate solutions with few observations). As an example, traditional priors like sparsity and more recently untrained network priors \cite{ulyanov2018deep} generalize better because they make fewer assumptions on the properties of the image manifold, but they suffer from poor quality of image reconstruction under few observations. In contrast, high quality generative priors have been shown to be highly efficient \cite{bora2017compressed,shah2018solving}, while being applicable only to images from the training distribution. Another issue with most existing priors, including generative priors, is that they require knowledge of the \emph{exact} measurement operator, which is unrealistic in a real world application where there are bound to be issues like sensor drift or environmental and hardware artifacts that may not be explained away by noisy observations. These are typically corrected by sensor calibration, requiring human intervention that is expensive and impractical. On the other hand, not accounting for these calibration artifacts can severely degrade the performance of existing techniques. 
 
\begin{figure*}[!t]
\centering
\includegraphics[width=0.9\linewidth,clip,trim=10 0 10 0]{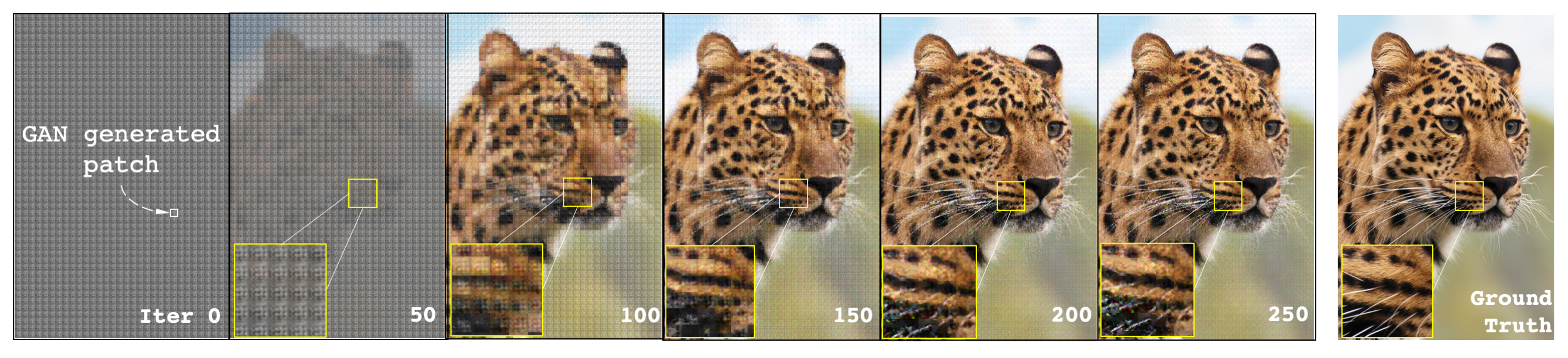}
  \vspace{-5pt}
 \caption{\small{\textbf{CS Recovery with GPP:} The generative patch prior is able to capture the diversity of the natural image manifold with a single pre-trained GAN irrespective of the size or aspect ratio, unlike existing generative priors that do not generalize outside of their training distribution. Here, we illustrate the recovery process (without any post processing) at a measurement rate of $10\%$. The original image is of size $1536\times1024$, and is recovered by a GAN trained on CIFAR-C \cite{hendrycks2019robustness} of size $32\times32$.}}
 \label{fig:color_example}
  \vspace{-15pt}
\end{figure*}
  
In this paper, we address these issues for the problem of compressive image recovery \cite{candes2008introduction,wakin2006architecture} using a deep generative \emph{patch} model that relaxes the prior from the image level to the patch level. We call this the Generative Patch Prior (GPP), and it exploits the relatively simpler parameterization of the patch manifold, instead of the image manifold, using a generative adversarial network. Since \name is defined at the patch level, it generalizes better than existing generative image priors, to a variety of natural images as well as images of arbitrary size and aspect ratio. In terms of the generalizability vs efficiency trade-off, GPP is significantly more generalizable (but less efficient) than a generative \emph{image} prior, while also being more efficient (but less generalizable) than existing unsupervised priors, making it much more practically applicable compared to both. Figure \ref{fig:color_example} demonstrates the recovery process using GPP on a previously unseen, high resolution image.

Within compressive imaging, the idea of using patch-manifolds can be further motivated by practical hardware acquisition conditions since measurement matrices for very high-resolution images can become impractically large. As a result, many practical compressive imagers are actually implemented as patch-level or block-compressive imagers  (c.f. \cite{kerviche2014information}). This means compressive measurements are indeed taken at the patch level, thus recovering the patches and then composing them into a picture requires a proper patch-manifold prior with a compositional prior. Further, practical CS imagers need an additional layer of calibration due to slight misalignments between the photosensor and the compressive optical elements. Usually, this calibration is done as a pre-processing step and fixed, but is prone to error and drift over time. To overcome this drawback, we additionally propose a self-calibrating (SC) mechanism using generative priors, i.e., a recovery process that is able to adjust automatically to the required calibration setting. We assume a scale and shift calibration model, which accounts for common calibration issues like sensor gain or other physical changes in the measurement process. We demonstrate empirically that GPP outperforms several unsupervised priors, including recent untrained network-based priors \cite{ulyanov2018deep,jagatap2019algorithmic}. 

\paragraph{Main contributions:}
\begin{enumerate}
\setlength\itemsep{-.1em}
    \item  We propose the generative patch prior (\name)-- which enhances the applicability of generative priors to the entire set of natural images, as opposed to only being restricted to the training distribution. 
    \item We show that the patch manifold can be approximated well by standard datasets (like CIFAR), for high quality CS image recovery, compared to existing unsupervised priors.
    \item We extend the idea of self-calibration for generative priors, which minimizes the need to intervene for manual sensor-level calibration in compressive imaging systems.
    \item Finally, we demonstrate that the proposed prior and calibration model achieves significantly better reconstruction quality on \emph{un-calibrated} measurements obtained from a real block compressive imager validating our approach under real unknown sensor noise and calibration.
\end{enumerate} 

\section{Background}

\label{sec:related_work}
\subsection{Preliminaries}
\vspace{-5pt}
Consider a vectorized square block of an image $\mathbf{x} \in \mathcal{X} \subset \mathbb{R}^n$ which we want to sense, and denote by $\mathbf{y} \in \mathbb{R}^m$ the compressive measurements obtained by the sensor. Given a measurement matrix $\Phi \in \mathbb{R}^{m \times n}$, with $m < n$ and $\Phi_{i,j} \sim \mathcal{N}(0,1)$, the compressive recovery problem is to estimate $\mathbf{x}$ accurately from $\mathbf{y}$. We consider three different sensing scenarios in this paper to benchmark the performance of the proposed patch-based prior--(a) Compressive sensing with known calibration (i.e. the ideal setting): $\mathbf{y} = \Phi \mathbf{x}$; (b) Compressive sensing with unknown calibration: $\mathbf{y} = (a \Phi + b\mathbf{1})\mathbf{x}$, where $a,b \in \mathbb{R}^1$ are unknown calibration parameters and have to be estimated, and $\mathbf{1} \in \mathbb{R}^{m\times n}$ is a matrix of the same size as $\Phi$ with $1$s; and finally (c) Compressive phase retrieval, which has non-linear forward process defined by $\mathbf{y} = |\Phi \mathbf{x}|$.

As $m < n$, reconstructing $\mathbf{x}$ from $\mathbf{y}$ in any of the above models is an ill-posed inverse problem and therefore has infinite number of feasible solutions. We define the Measurement Rate (MR) to be the ratio $\frac{m}{n}$. However, if we have additional knowledge about the structure of the solution space $\mathcal{X}$ (in this paper, implicitly via the set of natural patches), it is possible to recover $\mathbf{x}$ nearly perfectly given $m$ is sufficiently high and with graceful degradation as $m$ is reduced.
\vspace{-5pt}
\subsection{Related Work}
\vspace{-5pt}
\textbf{Priors for Inverse Problems:} Some of the most effective early work on designing priors for inverse problems came in the form of sparsity/compressibility in the wavelet and gradient domains \cite{candes2008introduction, candes2006robust,netrapalli2013phase,cai2016optimal,wang2017sparse} and hard-crafted prior models that were constructed based on such observations. Subsequently, more complex models were proposed such as wavelet trees and block sparsity \cite{baraniuk2010model,jagatap2017fast}, non-local low-rank regularization \cite{dong2014compressive} and denoiser-based regularization \cite{metzler2016denoising}. These models are ineffective at low measurement rates ($< 0.1$) and extremely slow in terms of computational complexity. For a given $\Phi$, supervised deep learning based methods have addressed these issues, like denoising autoencoders \cite{mousavi2015deep}, or convolutional neural networks (CNNs) \cite{kulkarni2016reconnet,boominathan2018phase,nguyen2018deep,rivenson2018phase}. These ideas have also been extended to the phase retrieval setting in applications such as Fourier Ptychography \cite{boominathan2018phase}. While these trained models do not work with new $\Phi$s straight away, there has been work to make them more adaptable \cite{lohit2018rate}. Another class of algorithms is obtained by unrolling/unfolding iterative algorithms which combines the advantages of the model-based and data-driven approaches, where a neural network serves as a sparsifying transform \cite{gregor2010learning,sun2016deep,zhang2018ista}, a learned denoiser \cite{metzler2017learned,metzler2018prdeep}, or a projector onto the image manifold \cite{rick2017one}. These methods overcome the drawbacks of earlier methods. However, they still rely on paired data for training. Some of these requirements are addressed by \cite{xia2019training}, where the image reconstruction algorithm is trained entirely using only the measurements for a given measurement operator. The limitation of paired training data can be overcome by using deep trained/untrained priors, which have the representational power of deep neural networks, while not requiring paired training like classical approaches. Trained priors like generative models \cite{bora2017compressed,shah2018solving,hand2018phase,shamshad2019deep}, and untrained network priors like deep image prior \cite{ulyanov2018deep,heckel2018deep,jagatap2019algorithmic} have been empirically shown to be useful compared to traditional priors for solving ill-posed problems in imaging. However, existing generative priors don't generalize out of distribution, and untrained network priors are extremely slow and fail under low sensing regimes; both of which are addressed by our proposed method, GPP. 

\textbf{Self Calibration:} In compressive sensing, the idea of self-calibration has been extensively studied as the problem of basis mismatch \cite{chi2011sensitivity,ling2015self}, although it is studied in the context of more traditional image priors such as sparsity in some spectral basis, whereas our focus is on more recent generative priors. 
More generally, the self-calibration problem has been studied in different contexts like medical imaging \cite{griswold2002generalized}, camera self-calibration \cite{pollefeys1999self} and in other bi-linear inverse problems \cite{ling2018self}. The SC problem is also indirectly related to several recent efforts in solving problems `blindly', with unknown corruptions \cite{anirudh2020mimicgan}, filters \cite{xia2019training,asim2019blind}, or demodulation \cite{hand2019global}. 
    

\textbf{Patch-based models:} Since smaller patches are statistically simpler and easier to model than entire image, priors designed from on image patches have been used in computer vision and image processing for a long time \cite{freeman2000learning,peyre2009manifold,ni2011example,cho2008patch}. GPP is also related to patch-based dictionary learning approaches used for denoising problems like KSVD \cite{elad2006image} where we expect the patch generator to act as a more powerful, non-linear basis than the one learned using KSVD. More recently, deep learning based on patches from a single image have emerged as powerful regularizers like in untrained network priors \cite{yokota2019manifold}, and generative models \cite{shaham2019singan}, and inverse imaging  \cite{Gilton2019}. These approaches do not focus on approximating the space of the entire patch manifold that can serve as a generic image prior, instead focusing on space of patches related to a single image. 

\section{A patch-manifold based generative prior for natural images}

While generative image priors have been reasonably successful in solving under-constrained inverse problems for low-resolution images ($128 \times 128$), they do not generalize to new domains that are not represented by the training distribution and are difficult to scale up to higher resolution natural images partly because the measurement operators for such images can become impractically large. Further, it remains extremely challenging to train a generative model to approximate the entire natural image manifold of complex, high resolution images. Instead, we propose a patch-based generative model which addresses these problems. As patches are assumed to be `small' and with much simpler spatial statistics than full images, the so-called ``patch manifold'' \cite{peyre2009manifold} is therefore easier to parameterize when compared to the true image manifold. This has been the driving factor behind using patch-based models in computer vision for the last two decades \cite{freeman2000learning,peyre2009manifold,cho2008patch}. 

In this paper, we are interested in using a generative adversarial network (GAN) \cite{GANGoodfellow} to approximate the patch manifold. As a result, we are able to leverage the representational power of GANs, and provide a \emph{differentiable} way to project onto the true image manifold. To the best of our knowledge, there does not exist such a patch-based generative prior for inverse problems. Given an image $\mathbf{x}\in \mathcal{X}$, the generative prior for images is implemented by constraining the solution to the inverse problem, $\mathbf{x}$ to the range space of a pre-trained generative model, $\{\mathcal{G}(\mathbf{z})\}~~ \forall~\mathbf{z}$, where the $\mathcal{G}$ is the generator, and $\mathbf{z} \in \mathbb{R}^d$ is sampled from a known probability distribution like the uniform distribution $\mathcal{U}(0,1)$. By definition, this generative image prior can only recover images that are expressed by the $\mathcal{G}$. Accordingly, the solution to the inverse problem is given by $\mathbf{x}^* = \mathcal{G}(\mathbf{z}^*)$, where for a loss function $\mathcal{L}$ like the mean squared error, $\mathbf{z}^* = \argmin_{\mathbf{z} \in [0,1]^d} \mathcal{L}(\mathbf{y},\Phi~\mathcal{G}(\mathbf{z}))$.

\textbf{Generative Patch Prior:} Formally, a patch is is defined as $x_p = \mathcal{F}(\mathbf{x},c,\tau)$, where $c$ is the center of the patch with width $\tau$, on an image $\mathbf{x}$ defined on uniform grid $[0,1]^2$, and $\mathcal{F}$ is the patch transform \cite{cho2008patch} which decomposes an image into $N_p$ equally sized non-overlapping patches. First, we train a GAN on a dataset of patches until convergence. This dataset can be obtained either by using the patch transform on a dataset with larger images like ImageNet \cite{russakovsky2015imagenet}, or directly using a dataset with smaller images like CIFAR \cite{krizhevsky2009learning}. While traditional patch-based methods assume much smaller patches (of size $4\times 4$ or $8\times 8$) to be texture-features with mostly low frequency spatial information, we find that datasets like CIFAR with relatively larger patches can form effective learned priors. Once we have the pre-trained GAN, the solution to the inverse problem with the generative image prior is reformulated in terms of a patch-based generator as follows: 

\begin{multline}
\mathbf{z}_i^* = \argmin_{\mathbf{z}_i \in [0,1]^d } \mathcal{L}(\mathbf{y}_i,\Phi~\mathcal{G}(\mathbf{z}_i)), \forall i = \{1, \dots, N_p\} \\ \mbox{~and~}\mathbf{x}^* = \mathcal{F}^{-1}(\mathcal{G}(\mathbf{z}^*_1),\dots,\mathcal{G}(\mathbf{z}^*_{N_p})),
\label{eq:gpp_prior}
\end{multline}

where $\mathbf{y}_i$ corresponds to the measurements obtained under the measurement operator $\Phi$ for each individual patch, and $\mathcal{F}^{-1}: \mathcal{P}\times \mathcal{P}\dots \times \mathcal{P} \mapsto \mathcal{X}$ is the inverse patch transform, which we also refer to as a compositional prior, and takes a set of patches to produce the final image. Since we treat the patches independently, under very few measurements this can lead to blocking artifacts. This can be alleviated using a filtering mechanism like BM3D \cite{dabov2006image}. For fairness, we compare all baselines with the same filtering on the recovered image. Figure \ref{fig:color_example} shows the optimization process for a high resolution image of size $1536\times1024$, of non-standard aspect ratio recovered with a measurement rate of $10\%$. 

\paragraph{Relationship to generative \emph{image} priors:} GPP relaxes the generative image prior to the level of a patch, and as a result is far more applicable. The GPP can effectively recover any natural image with nearly zero representation error unlike GAN priors. The two become equivalent when the size of the patch is equal to the size of the image, i.e., $N_p=1$.  A generic unconditional GAN that approximates the natural image manifold does not exist today, however, in cases where GANs can be trained well like faces, animals, cars etc. it is expected that for an image from that specific distribution, the GAN prior will outperform GPP at low measurement rates. We explore these properties further in the experiments section.

\section{Self Calibration with \name}
\label{sec:gpp_sc}
Most existing deep prior formulations in literature assume access to the exact measurement matrix $\Phi$ in order to solve the inverse problem. Any changes to $\Phi$ that are unknown can render such techniques useless due to errors in the optimization objective. This creates practical issues since there are likely going to be differences in the measurement operator realized using hardware sensors compared with the operator used during training. The exact nature of change in the hardware is hard to determine in advance and this issue is corrected by a manual calibration step requiring human intervention. Instead, here we pursue a self-calibrating approach that reduces the need for such manual interventions.

\textbf{Calibration model:} We propose a form of self-calibration that relaxes the need to know the exact $\Phi$. We assume a simple calibration model using two parameters: $\Tilde{\Phi} = \tilde{a}(\Phi + \tilde{b}\mathbf{1})$, where $\tilde{a} \in \mathbb{R}$ accounts for changes in sensor gain (mapping maximum intensity color white to 1.0), $\tilde{b} \in \mathbb{R}$ models unknown parameters in the measurement operator with a simple bias term and $\mathbf{1} \in \mathbb{R}^{m \times n}$ denotes a matrix of ones. For notational convenience, we rewrite the calibration model as $\Tilde{\Phi} = a\Phi + b\mathbf{1}$. The compressive sensing problem is re-formulated as $\mathbf{y} =  (a\Phi + b\mathbf{1})\mathbf{x}$, with three unknowns: $a, b, \mathbf{x}$. For known $a,b$, i.e. a manually calibrated measurement, $\mathbf{x}$ can be readily estimated using \eqref{eq:gpp_prior} using the proposed \name prior, or any other image-prior. Note that the loss function for patch $i$ now becomes:

\begin{equation}
\label{eq:calibration_cost}
\mathcal{L} = \norm{ \mathbf{y}_i-(a\Phi+b\mathbf{1}) \mathbf{x}_i) }^2 
\end{equation}

where $\mathbf{x}_i = \mathcal{G}(\mathbf{z}_i)$ for some $\mathbf{z}_i$.

\textbf{Solving for $a$ and $b$:} We use alternating minimization to solve for the unknowns $a,b,\mathbf{x}^*$. First, given an $\mathbf{x}$ from the \name model we estimate $a,b$ assuming the current estimate of $\mathbf{x}$ is the true solution. As our calibration model is linear in $a,b$ these can be solved exactly to minimize the cost in Eqn. \eqref{eq:calibration_cost}:

\begin{equation}
\label{eq:a_and_b_star}
a^* = \frac{c_\mathbf{1}\lambda-c_{\Phi}\theta_\mathbf{1}}{\lambda^2-\theta_\Phi\theta_\mathbf{1}} \mbox{~and~}b^* = \frac{c_{\mathbf{1}}-a^*\lambda}{\theta_{\mathbf{1}}},
\end{equation}

where we define scalar quantities for notational convenience: $c_{\Phi} = \mathbf{y}^\T\Phi\mathbf{x}, c_{\mathbf{1}} = \mathbf{y}^\T\mathbf{1}\mathbf{x},  \theta_{\Phi} = (\Phi\mathbf{x})^\T(\Phi\mathbf{x}), \theta_{\mathbf{1}} = (\mathbf{1}\mathbf{x})^\T(\mathbf{1}\mathbf{x}), \lambda = (\Phi\mathbf{x})^\T(\mathbf{1}\mathbf{x})$. It is easy to check that, in the ideal case when $\mathbf{y} = \Phi \mathbf{x}$ (i.e. no calibration is necessary), $c_\mathbf{1} = \lambda; c_\Phi = \theta_\Phi$ and as a result we obtain $a^* = 1, b^* = 0$ as expected. The details of complete derivation are available in the appendix. The algorithm for self-calibration with \name is shown in Algorithm \ref{alg:main}.

\begin{algorithm}[!htb]
\small{
\begin{algorithmic}[1]
\caption{Self-calibration with the generative patch prior}
\label{alg:main}
\Input{Patch-wise observations $\mathbf{y_i}$, $i = 1\dots N_p$, Measurement operator $\Phi$, Pre-trained patch generator $\mathcal{G}$.}
\Output{Recovered full Image $\mathbf{x}^* \in \mathbb{R}^{n N_p}$, calibration parameters $a^*,b^*$.}
\Input{Initialize: $\mathbf{z}^{(0)}_i \sim \mathcal{U}(\mathbf{0},I)~ \mbox{for patches~} i = 1 \dots N_p$}
\While{not converged}
    \State {$\mathbf{x}^{(t)}_i = \mathcal{G}(\mathbf{z}_i) ~\forall~ i = 1, \dots, N_p$}
    \State {Given $\mathbf{x}^{(t)}_i$, compute $a_i^{(t+1)}, b_i^{(t+1)}$ from \eqref{eq:a_and_b_star} $\forall i$}
    \State {$a^{(t+1)} \leftarrow \frac{1}{N_p}\sum^{N_p}_i a_i^{(t+1)};\mbox{ and } b^{(t+1)} \leftarrow \frac{1}{N_p}\sum^{N_p}_i b_i^{(t+1)}$}
    \State Compute cost $\mathcal{L}$ from \eqref{eq:calibration_cost}
    \State Latent space optimization: $\mathbf{z}^{(t+1)}_i\leftarrow \mathbf{z}^{(t)}_i-\gamma_g~\nabla_{\mathbf{z}}\mathcal{L} \quad \forall i$
    \EndWhile
 \State $\mathbf{x}^* = \mathcal{F}^{-1}(\mathcal{G}(\mathbf{z}^*_1),\dots,\mathcal{G}(\mathbf{z}^*_{N_p}))$
\end{algorithmic}
}
\end{algorithm}

Since our generative model is defined at a patch level, we estimate $a,b$ for each individual patch $i$ separately and assign the mean values of all the patch-estimates as the single $a,b$ for the entire image: $a = \frac{1}{N}\sum_{i=1}^N a_i$; and $b = \frac{1}{N}\sum_{i=1}^N b_i$. Finally, by plugging in the estimates for $a^*, b^*$ in the cost function \eqref{eq:calibration_cost}, we can solve for $\mathbf{x}^*$ using the optimization outlined in \eqref{eq:gpp_prior}. We continue with this alternating minimization until the loss in \eqref{eq:calibration_cost} converges. In practice, we find that the algorithm converges within 1500 iterations, and finding recovery and convergence properties of this algorithm remain part of our future work.


\section{Experiments}
\label{sec:experiments}
In this section, we benchmark the performance of the proposed \name and other commonly used unsupervised priors for compressive image recovery. We show that \name serves as a highly effective prior under very low sensing scenarios ($1-10\%$ measurement). We also demonstrate the robustness of the proposed self-calibration by introducing artificial calibration errors in the form of gain ($a$) and shift ($b$). Finally, we demonstrate that the combined model \name with self-calibration can recover very high quality images directly from un-calibrated measurements obtained using a real compressive imaging camera \cite{kulkarni2016reconnet}.
\paragraph{Deep generative patch model training} In our experiments we use non-overlapping patches of size of $32\times32$, and train a DCGAN \cite{radford2015unsupervised} for 100K iterations in TensorFlow. We experiment with several different datasets to approximate the patch manifold---tinyimagenet \footnote{https://tiny-imagenet.herokuapp.com/}, Imagenette\footnote{https://github.com/fastai/imagenette}, CIFAR\cite{krizhevsky2009learning}, CIFAR-C \cite{hendrycks2019robustness}, and MIT Places \cite{zhou2014learning}. In all the following experiments we report results from the GAN trained on CIFAR-C as the default.

\begin{table*}[!htb]
\centering
\small{
		\begin{tabular}{p{1.0in}p{0.35in}p{0.35in}p{0.35in}p{0.45in}p{0.45in}p{0.35in}p{0.42in}|p{0.35in}}
			\hline
			\textbf{Method}& 	\textbf{Barbara} & 	\textbf{Parrot} & 	\textbf{Lena} & 	\textbf{Foreman} & \scriptsize{\textbf{Cameraman}} & 	\textbf{House}& 	\textbf{Monarch} & \textbf{Avg.} \\ \hline
			  \multicolumn{8}{c}{Measurement Rate 10\%}\\
			& & &  && & &\\
			TVAL3 \cite{li2013efficient} &\textbf{22.21} &23.16 & \uline{24.16} & \uline{28.74} & \textbf{21.92} & 26.32&\uline{21.16} & \uline{23.95} \\
			NLR-CS \cite{dong2014compressive} &14.67 &14.16 & 15.33 &13.56&14.22 & 14.80&14.67&14.49 \\
			D-AMP \cite{metzler2016denoising} &21.23 &21.64 & 22.47 &25.58&20.35 & 24.71&19.00&22.14\\
			DIP \cite{ulyanov2018deep,jagatap2019algorithmic} & 21.20 & 22.88 & 23.02 &26.92 & 21.74 & 24.47& 20.68 & 22.98\\
			\name +SC (ours) &22.18 & \uline{23.27}&21.95 &28.11 &21.18 &\uline{26.41} &19.46 & 23.22\\
			\name (ours) & \uline{22.19} & \textbf{23.78} & \textbf{24.21} & \textbf{29.05} &\uline{21.61} &\textbf{26.99} &\textbf{21.72} & \textbf{24.22}\\
			\\\hdashline
			
			  \multicolumn{8}{c}{Measurement Rate 1\%}\\
			& & &  && & &\\
			TVAL3  &11.96 &11.46 &11.89&11.01 &12.00 &11.90 &11.11& 11.61\\
			NLR-CS & 5.86&5.44& 6.27 &4.25 & 6.31&5.29&6.71 & 5.73\\
			D-AMP  & 5.48&5.09 & 5.96 &3.83 & 5.64&5.02&6.20 & 5.31\\

			DIP & 14.07 & 12.83 & 14.68 &13.53 & 12.42 & 14.89& 13.64 & 13.72\\
			\name+SC (ours) & \uline{17.16}& \uline{17.09}&\uline{16.81} &\uline{18.30} &\uline{15.62} &\uline{16.65} &\uline{14.57} & \uline{16.59}\\
			\name(ours) & \textbf{17.45}& \textbf{17.46} & \textbf{17.31} & \textbf{20.40} & \textbf{16.63} &\textbf{18.89}&\textbf{14.58} & \textbf{17.53}\\
		

		\end{tabular}
		}
		\caption{\small{PSNR (dB) for compressive image recovery under ideal operating conditions, i.e. the exact measurement operator $\Phi$ is known. We observe that GPP performs favourably compared to several commonly used unsupervised priors, particularly with extremely few ($1\%$) measurements. Best performing method is shown in \textbf{bold}, and the second best is \uline{underlined}.}}
\label{tab:linear_cs}
\end{table*}
\subsection{Compressive Image recovery with \name}
We first study the problem of compressive sensing, where we are provided with $n$ measurements of an unknown image obtained as a linear projection of a measurement operator $\Phi$, which is typically a Gaussian random matrix. Our test image set consists of 7 standard images used commonly to benchmark compressive sensing reconstruction algorithms. These images are grayscale, and of size $256\times256$, therefore using non-overlapping patches of size $32\times 32$, every image is represented as a batch of $\frac{256^2}{32^2} = 64$ patches. The measurements are taken by vectorizing each patch so our $n = {32^2} = 1024$, and $\Phi \in \mathrm{R}^{m\times1024}$. We perform reconstruction using the two methods proposed here: (i) \name and (ii) \name + SC, which also includes the self-calibration step in addition to projected gradient descent (PGD) over the range of the patch-generator. We study the performance in two settings -- $1\%$ measurement rate, where $m = 10$, and $10\%$ measurement rate with $m = 102$. We use the Adam optimizer \cite{kingma2014adam} in all our experiments. For \name, we use a learning rate (LR) of $5e-3$, and run PGD for 1500 steps; for \name+SC we use LR = $5e-2$ for the first 200 iterations, and LR = $5e-3$ for the next 1800.
\begin{figure}
  \begin{center}
    \includegraphics[width=0.75\linewidth,clip,trim=5 35 5 40]{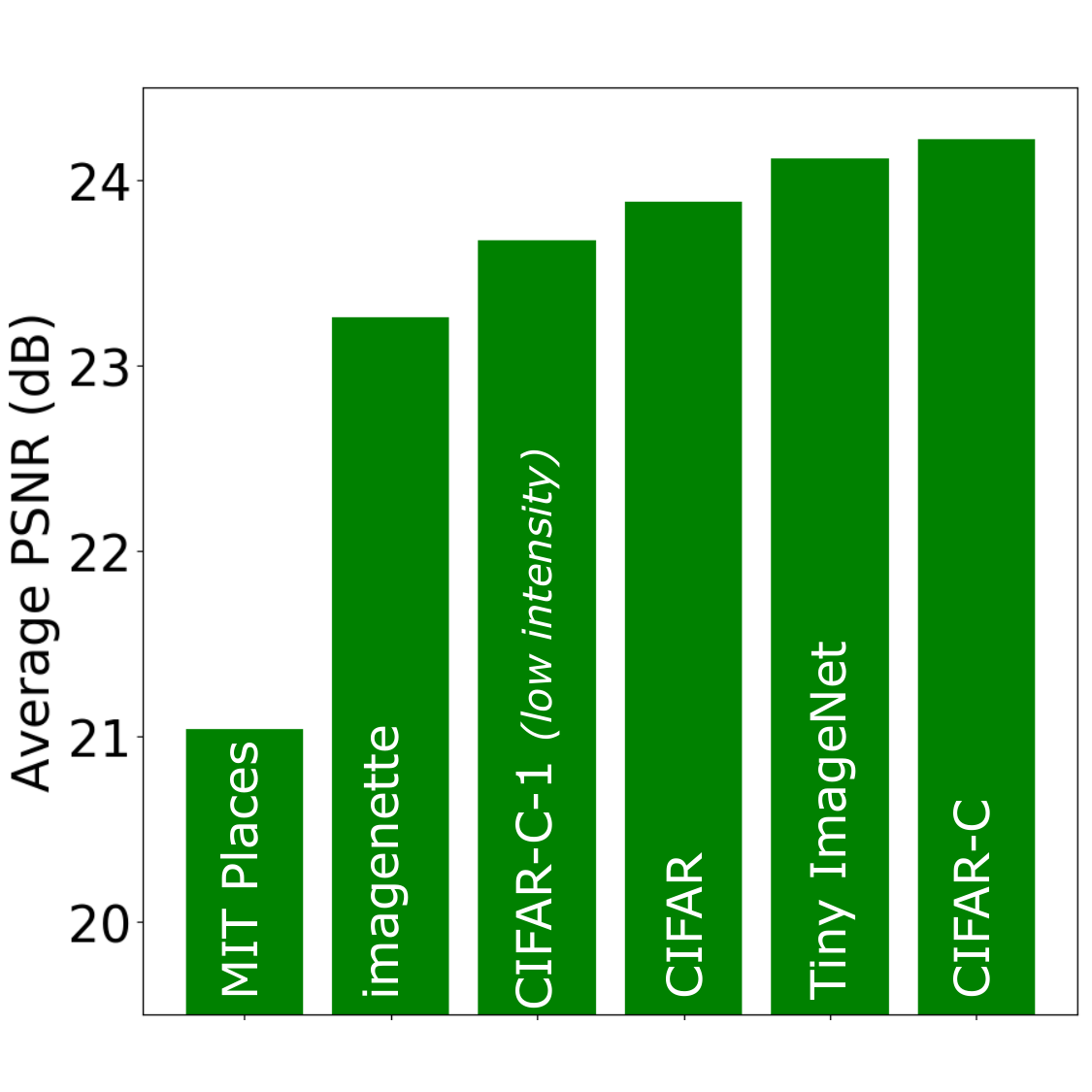}
  \end{center}
  \caption{\small{Approximating the patch-manifold with commonly used image datasets. Here we train a patch-generator with each dataset shown and use it as the backend for GPP. We evaluate it's effectiveness in the compressive image recovery problem at $10\%$ measurement rate and report average PSNR values.}}
     \label{fig:patch_dataset}
     \vspace{-20pt}
\end{figure}

\noindent \textbf{Baselines:} In addition to popular unsupervised benchmarks such as TVAL3\cite{li2013efficient}, NLR-CS\cite{dong2014compressive}, and D-AMP\cite{metzler2016denoising}, we implement a patch version of deep image prior (DIP) \cite{ulyanov2018deep} or untrained network priors \cite{jagatap2019algorithmic}. Here, we train a randomly initialized network to predict a batch of image patches, instead of a single image. We do this because it becomes impractical to implement a dense measurement operator for the full image of size $256\times 256$, and we also observe that the patch-wise version converges much quicker. We use the same patch wise for comparison as in \name with a LR = $5e-4$ and run it for 10000 iterations for each image.

\noindent \textbf{Results:} We report the results for $1\%$ and $10\%$ measurement scenarios in table \ref{tab:linear_cs}. In all the methods, we report PSNR after post-processing with BM3D \cite{dabov2006image}. For DIP, \name, and \name+SC we use $\sigma=0.25$ for the $10\%$ measurement and $\sigma=0.5$ for the $1\%$ scenario, for the other baselines we report numbers on $33\times33$ sized patches from \cite{kulkarni2016reconnet}. We see that the two proposed methods, \name and \name+SC perform very similarly. As expected, using self-calibration in addition to PGD optimization suffers a little when operating in ideal conditions. Yet, \name+SC performs on par with many unsupervised baseline methods. Next, we observe that in both the cases \name outperforms all the competing unsupervised methods. It is particularly noteworthy that for the $10\%$ measurement case, the performance of \name is higher than patch-based \emph{supervised} methods like ReconNet \cite{kulkarni2016reconnet} and SDA \cite{mousavi2015deep} (not shown in table) which have an average PSNR of 24.09dB and 23.49dB, respectively. In the $1\%$ measurement case in table \ref{tab:linear_cs}, we observe that \name and \name+SC significantly outperform the state-of-the-art unsupervised baselines considered by nearly $4dB$ indicating the effectiveness of patch-based priors in extreme sensing scenarios. An ablation on the choice of dataset to construct \name is shown in figure \ref{fig:patch_dataset}. We observe that CIFAR/ImageNet datasets form better priors when compared to MIT Places, perhaps because MIT Places mostly contains low frequency, smooth patches of natural locations, whereas CIFAR contains a lot more high frequency spatial information within each patch.
\begin{figure*}[!t]
\vspace{-10pt}
\centering
\includegraphics[width=0.95\linewidth,clip,trim=0 0 0 0]{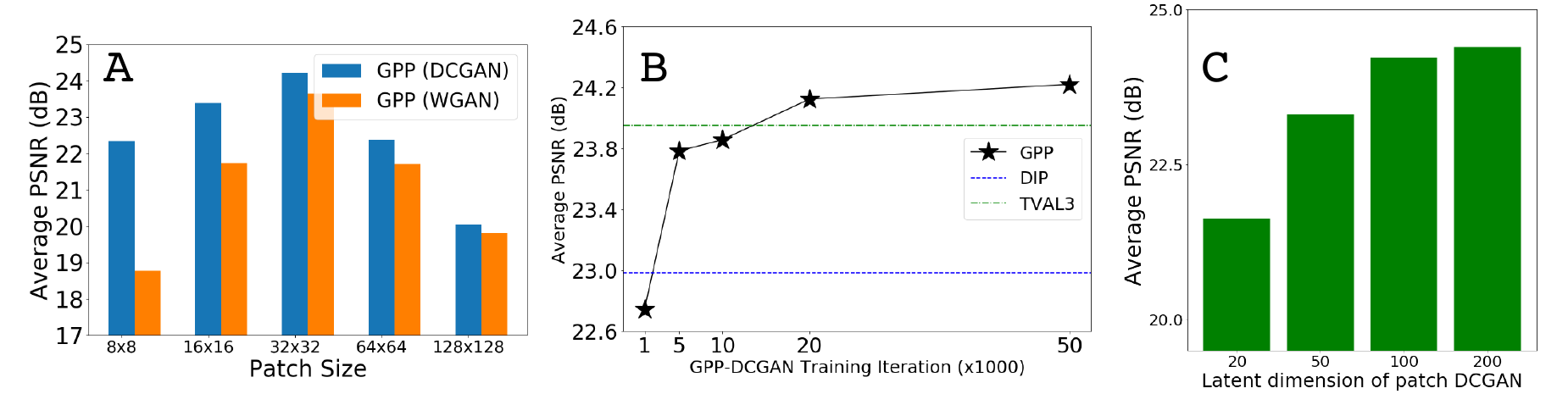}
  \vspace{-10pt}
  \caption{\small{\textbf{(A)} Ablation on Patch Size in GPP. PSNR (dB) is shown across two GAN variants DCGAN and Wasserstein GAN (WGAN). \textbf{(B)} Comparing quality of reconstruction vs GAN training iterations. We see that even with a partially trained GAN backbone, GPP provides a benefit over the next best performing methods. (\textbf{C}) Ablating the affect of latent dimension of the patch DCGAN. }}
  \label{fig:ablation}
\end{figure*}

\begin{figure*}[!htb]
\centering
\includegraphics[width=0.99\linewidth,clip,trim=10 0 20 0]{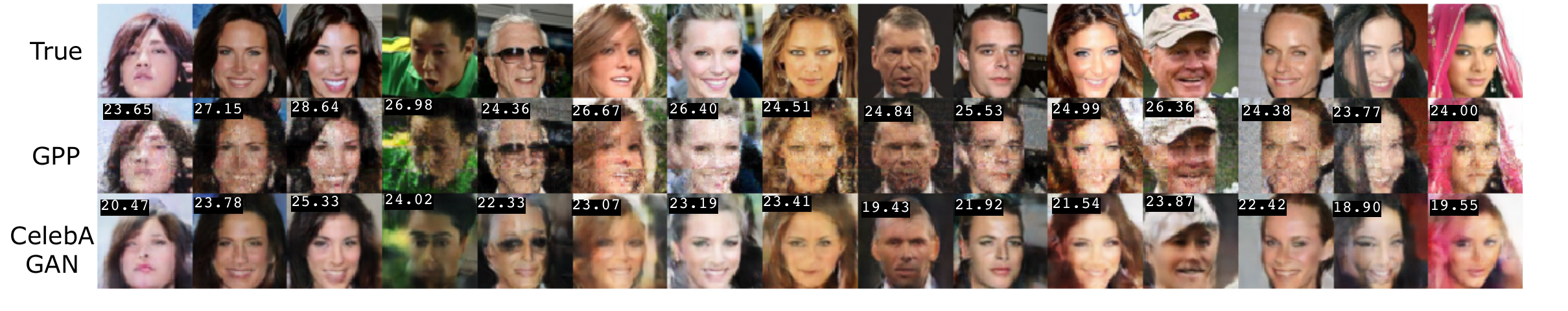}
  \caption{\small{Comparing GPP's performance on CelebA faces with a DCGAN trained on CelebA Faces for compressive image recovery with $50\%$ measurement rate. The GAN prior is effective when low number of measurements are available, but only when the true solution is from the same distribution on which the GAN is trained. On the other hand, GPP applies to any natural image and with more measurements, outperforms the GAN prior because it is able to recover the true image with a much lower projection error. See table \ref{tab:celebA} for quantitative comparisons. Here we use the same GPP model as before, trained using CIFAR-C images with patches of size $16\times 16$ to recover CelebA faces.}}
  \label{fig:celeba}
\end{figure*}
\subsection{Properties of GPP}
We examine the components of GPP in order to better understand GPP's behaviour. (a) \textbf{Patch sizes:} First, we study the effect of the size of the patches -- in a patch sensing setup, as the size of patches become smaller, the number of measurements available in each patch become correspondingly fewer. On the other hand, a larger patch implies fewer patches to compose an image, thereby limiting the details that can be resolved. In the extreme, when the size of the patch is equal to the size of the image, the GPP becomes equivalent to generative priors that have been used recently in inverse problems. In Figure \ref{fig:ablation}(A), we show the average PSNR with varying patch size, as expected we see that a patch size of $32$ seems to be optimal, with performance degradations with increasing or decreasing sizes. (b) \textbf{Effect of training DCGAN:} Next, we study the importance of training the DCGAN in figure \ref{fig:ablation}(B), where we see that even a partially trained patch-generator can give an improvement over standard baselines. (c) \textbf{Latent dimension of the patch generator:} Finally, in figure \ref{fig:ablation}(C) we show how the performance varies with varying dimensionality of the patch-DCGAN. As expected, we observe quality increases with an increased capacity of the generator, indicating that better patch-generators can further improve image recovery, however this comes at a cost of more complexity in training the generator. In all our experiments we use a patch size of $32$, and a latent dimensionality of $d = 100$.

\noindent \textbf{GPP and generative image priors}
As noted earlier, GPP is related to the GAN prior when the number of patches used $N_p=1$. Here, we study how this affects the performance in compressive image recovery. Since GAN priors, by design, are restricted in their use on specific datasets on which they are trained, we perform this experiment on the commonly used CelebA Faces dataset \cite{liu2015faceattributes}. Here we train a standard DCGAN on a training set and perform CS recovery on 100 images from the test set using different number of measurements. We also use the same GPP as before, trained on CIFAR-C images as a comparison on the same CelebA test images. These results are reported in table \ref{tab:celebA} and we remark on a few observations. First, since these images are faces we report both PSNR and SSIM evaluation metrics. Next, as expected, we find that under very few observations the Image-level generative prior works better simply because it can find a similar face, whereas GPP cannot. On the otherhand, once more observations are available, GPP outperforms the GAN prior on both metrics considered here because it is able to find the \emph{exact} face better than the GAN prior. A few recovered examples are shown in Figure \ref{fig:celeba}.

\begin{table}[!htb]
\centering
\small{
\begin{tabular}{p{0.25in}p{1.0in}p{0.6in}p{0.5in}}
			\hline
			\textbf{Meas}& 	\textbf{GAN Prior} & 	\textbf{PSNR} (dB) & 	\textbf{SSIM} \\ \hline
			\multirow{2}{*}{$5\%$} & Image-level \cite{bora2017compressed,shah2018solving} & \textbf{19.170} & \textbf{0.4752}\\
									& Patch-level \emph{(ours)} & 16.770 & 0.2597\\
				\hline
				\multirow{2}{*}{$15\%$} & Image-level & \textbf{21.367} & \textbf{0.5826} \\
										& Patch-level & 20.663 & 0.5294\\
						\hline
				\multirow{2}{*}{$50\%$} & Image-level & 22.913 & 0.6478 \\
				& Patch-level & \textbf{26.813} & \textbf{0.8261}\\
						\hline

	\end{tabular}
	}
		\caption{\small{CS reconstruction on the 100 test images from the CelebA Faces dataset \cite{liu2015faceattributes}. We compare our patch-based prior trained on CIFAR-C with an image-based prior which is a DCGAN trained on CelebA. As expected, under very few observations the image prior is better because it can find a similar face but this only works when the images are from the same distribution on which the GAN is trained. Yet with more observations GPP outperforms even the GAN prior because it is able to find the exact face.}}
\label{tab:celebA}
\end{table}
\subsection{Self-Calibrated image recovery from a physical compressive camera}
\vspace{-5pt}
Here we study a real world use case of the self-calibrating mechanism in recovering high quality images from patch-based measurements that obtained using a real world compressive imaging system \cite{kerviche2014information}. We use the measurements collected under a measurement rate of $10\%$, provided by the authors of \cite{kulkarni2016reconnet}, \cite{lohit2018convolutional}. Following \cite{kulkarni2016reconnet}, the patches are $33\times33$, and so we resize the output of the generator from $32\times32$ to match this patch size. As expected, this setup requires manual calibration in order to use existing algorithms out of the box, without which they produce poor reconstructions. We show results over 7 different \emph{un-calibrated} measurements and their corresponding reconstructions in table \ref{tab:real_dataset}. In order to compute PSNR, we use the recovered image from every algorithm and resize it to $256\times256$ so it is comparable to the original, and scale it to have a maximum intensity of 1.0, which itself acts as a simple calibration for the baselines. This normalized image is compared to the ground truth to evaluate fidelity. Finally, we process the reconstructions from all the methods using BM3D \cite{dabov2006image}, using $\sigma=0.25$. In addition to the baselines described previously, we also report results from ISTA-Net\cite{zhang2018ista} which is a supervised, state-of-the-art CS reconstruction technique that also uses $33\times33$ patches. We also add a "manual calibration" (MC) setting with \name, where we use these estimates for $a$ and $b$ obtained using the \name+SC, and use only \name (without SC) as if these calibration parameters were known. Finally, for this setting we train DIP for 50000 steps before observing convergence, for both the un-calibrated and manually calibrated settings.

\begin{table*}[!htb]
\centering
\small{
\begin{tabular}{p{1.0in}p{0.35in}p{0.35in}p{0.35in}p{0.45in}p{0.45in}p{0.35in}p{0.4in}|p{0.35in}}
			\hline
			\textbf{Method}& 	\textbf{Barbara} & 	\textbf{Parrot} & 	\textbf{Lena} & 	\textbf{Foreman} & \scriptsize{\textbf{Cameraman}} & 	\textbf{House}& 	\textbf{Boats} & \textbf{Avg.} \\ \hline
			TVAL3 \cite{li2013efficient} &17.65 & 12.55 & 18.05 &13.16 & 15.36 &14.29 &14.82 & 15.13\\
			DIP \cite{ulyanov2018deep,jagatap2019algorithmic} &6.75 & 7.87 & 9.13 & 4.52 & 8.22 &6.15 &6.06 & 6.96\\
			ISTA-Net\cite{zhang2018ista} &\textbf{18.95} & 16.23 & 19.19 &15.94 & \textbf{17.11} &18.01 &16.14 & 17.37\\
			\name (ours)                & 12.78 & 12.66 & 13.50 & 10.73 & 13.32 &12.77 &11.83 &12.54\\
			DIP+SC      & 17.97 & 18.46 & 15.86 & 20.51 & 19.58 &17.64 &18.66 &18.39\\
			\name+SC (ours)        & 18.21 & \textbf{20.01} &\textbf{20.32} &\textbf{20.68} &16.33 &\textbf{20.47} &\textbf{18.29} & \textbf{19.20}\\\hdashline
			DIP+\footnotesize{Manual calib.}        & 18.88 & 19.67 & 19.97 & 19.59 &19.64 &19.74 &17.57 & 19.29\\
			\name+\footnotesize{Manual calib.}        & 19.06 & 20.23 & 20.43 & 19.97 &19.52 &20.22 &17.87 & 19.61\\\hline

	\end{tabular}
	}
		\caption{\small{\textbf{Real CS dataset:} PSNR (dB) is shown for 7 test images obtained using \emph{real, un-calibrated} measurements \cite{lohit2018convolutional} from a patch-based compressive camera at MR = $10\%$. The proposed self-calibration improves the quality of recovered images significantly compared with baselines that require manual calibration.}}
\label{tab:real_dataset}
\end{table*}
\begin{figure*}[!htb]
\centering
\includegraphics[width=0.9\linewidth,clip,trim=0 0 0 0]{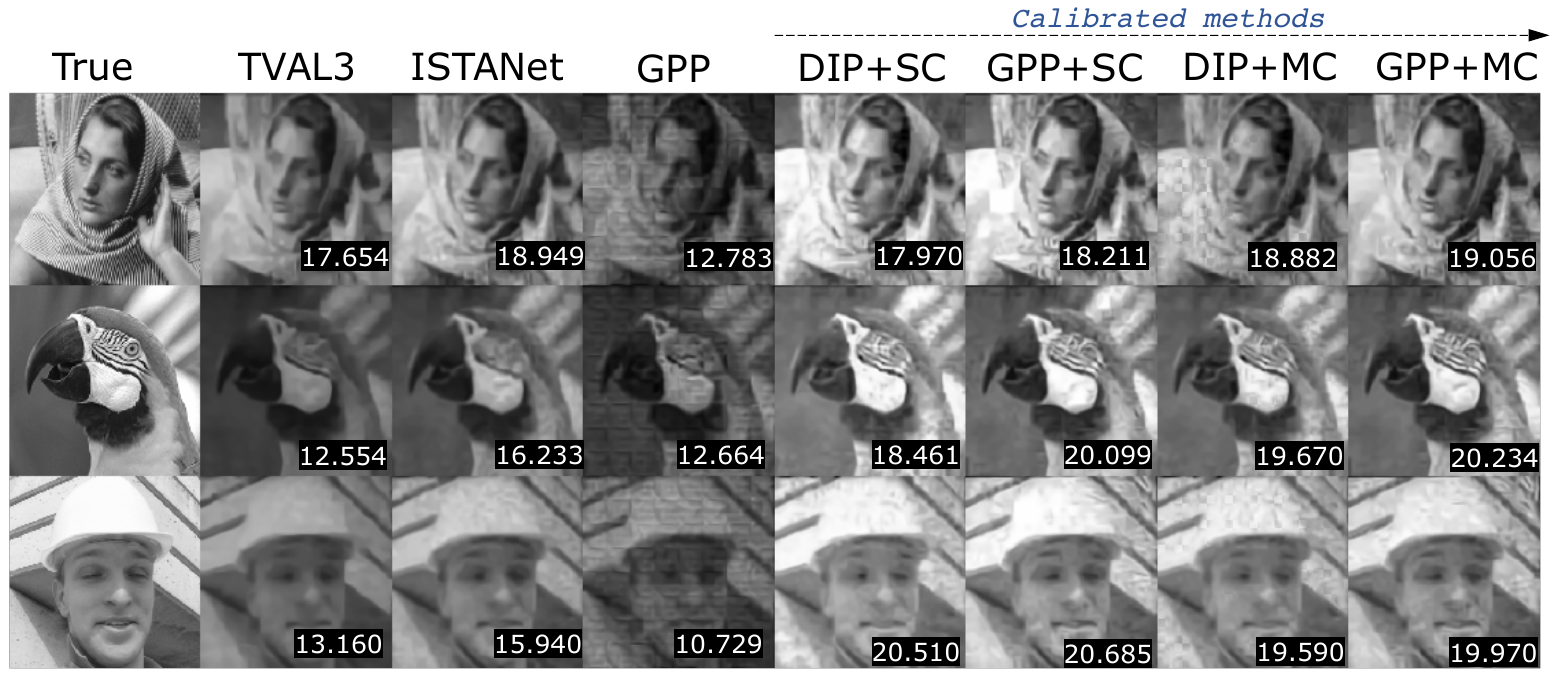}

 \caption{Sample reconstructions on un-calibrated measurements from a real compressive imaging system. PSNR (dB) is shown at the bottom right of each image. MC refers to manual calibration, using parameters obtained using \name+SC. For the real data, we estimate the unknown calibration parameters to be around $a=0.08,~b = 0.0$. We don't show DIP results here because it performs very poorly compared to other baselines.}
 \label{fig:sample_recon}

\end{figure*}
\vspace{5pt}

\noindent \textbf{Results} Table \ref{tab:real_dataset} shows performance in terms of PSNR. In these experiments, we estimate the calibration parameters to be $a = 0.075 - 0.085;~b = 0.0$. The sensor gain $a$ varies for each image, which is expected since every image may require slightly different calibration settings. In the manual calibration setting, we use an average value of $a = 0.08,~b = 0.0$.  We find that \name+SC is able to successfully recover a high quality image from the un-calibrated measurements, unlike all baseline approaches, resulting in a gain of ~4dB over the next best unsupervised method. We also observe that \name+SC provides better reconstructions when compared to ISTA-Net \cite{zhang2018ista}, which is a supervised technique. For reference, we show the manual calibration settings for both \name, and DIP, and observe that they are only marginally better than \name+SC. A few qualitative comparisons are shown in figure \ref{fig:sample_recon}.

\subsection{Compressive Phase Retrieval}
\begin{table}
\centering
\begin{tabular}{cc}
			\hline
			\textbf{Method}& \textbf{Avg. PSNR (dB)} \\ \hline

			DIP \cite{ulyanov2018deep,jagatap2019algorithmic}& 18.439\\
			\name (ours) & \textbf{22.616}\\
	\end{tabular}
	\caption{Phase Retrieval at a measurement rate of $10\%$. }
	\label{tab:phase_retrieval}

\end{table}

Finally, we evaluate \name on the non-linear inverse problem of phase recovery in compressive sensing. We use the same experimental settings as in the linear compressive sensing case. We implement the alternating phase projected gradient descent described (APPGD) in \cite{hyder2019alternating}, which is shown to have better performance compared to other gradient descent based optimization methods. We compare against the deep image prior (DIP) method, recently used in \cite{jagatap2019algorithmic} for phase retrieval and show favourable performance for the $10\%$ measurement case shown in table \ref{tab:phase_retrieval}.

\section{Discussion}
We presented a new kind of deep prior called the generative patch prior (GPP) which uses a GAN trained on patches instead of full sized images as a learned prior for inverse problems. Using GPP, we address the problem of compressive image recovery and show that GPP has many favourable properties compared to existing generative priors -- it is more broadly applicable to a wide variety of images not seen during training, or images or arbitrary sizes and aspect ratios. We also see that GPP outperforms several commonly used learned and model priors in compressive sensing and compressive phase retrieval tasks. We also propose a self-calibration mechanism that enables the model to estimate and adjust to calibration artifacts automatically. Finally, we validate this mechanism and the GPP on a real, un-calibrated compressive sensing dataset and demonstrate that the proposed performs well without requiring manual calibration unlike existing methods.

\newpage
\section*{Broader Impact}
\textbf{Applications:} This work applies to the compressive sensing (CS) problem, which has been shown to result in faster, cheaper, bandwidth and energy-efficient sensing. As a result, CS methodologies have found applications in a diverse set of applications ranging from MRI, to privacy-preserving sensing, and even for energy efficient visual analytics. While the primary focus of this work is in advancing algorithmic aspects which are broadly applicable, it is applicable to the full spectrum of problems by a simultaneous effort in domain-specific hardware advances.

\textbf{Implications:} As indicated above, the societal implications are strongly tied to the end-application integrated with the specifics of the hardware. There is already existing work on using similar techniques in faster MRI in the healthcare sector. Security and surveillance technology stands to benefit by making analytics robust to under-sampling of visual information; making cameras portable etc. We acknowledge that these approaches, if pushed further, can expand the surveillance capabilities to the point that they may open many ethical questions. While we don't have a solution to prevent misuse, we are aware of its potential. We are, however, motivated by its significant potential for applications in health, basic scientific inquiry, and more broadly energy efficient sensing.

\bibliographystyle{ieee}
\bibliography{references}
\section*{Acknowledgements}
The first author's work was performed under the auspices of the U.S. Department of Energy by Lawrence Livermore National Laboratory under Contract DE-AC52-07NA27344.
\section*{Disclaimer}
 \noindent This document was prepared as an account of work partly sponsored by an agency of the United States government. Neither the United States government nor Lawrence Livermore National Security, LLC, nor any of their employees makes any warranty, expressed or implied, or assumes any legal liability or responsibility for the accuracy, completeness, or usefulness of any information, apparatus, product, or process disclosed, or represents that its use would not infringe privately owned rights. Reference herein to any specific commercial product, process, or service by trade name, trademark, manufacturer, or otherwise does not necessarily constitute or imply its endorsement, recommendation, or favoring by the United States government or Lawrence Livermore National Security, LLC. The views and opinions of authors expressed herein do not necessarily state or reflect those of the United States government or Lawrence Livermore National Security, LLC, and shall not be used for advertising or product endorsement purposes.
\onecolumn
\appendix
\begin{appendices}

  \section{Additional results}
  \begin{figure*}[!htb]
  \centering
  \includegraphics[width=0.99\linewidth,clip,trim=0 30 0 0]{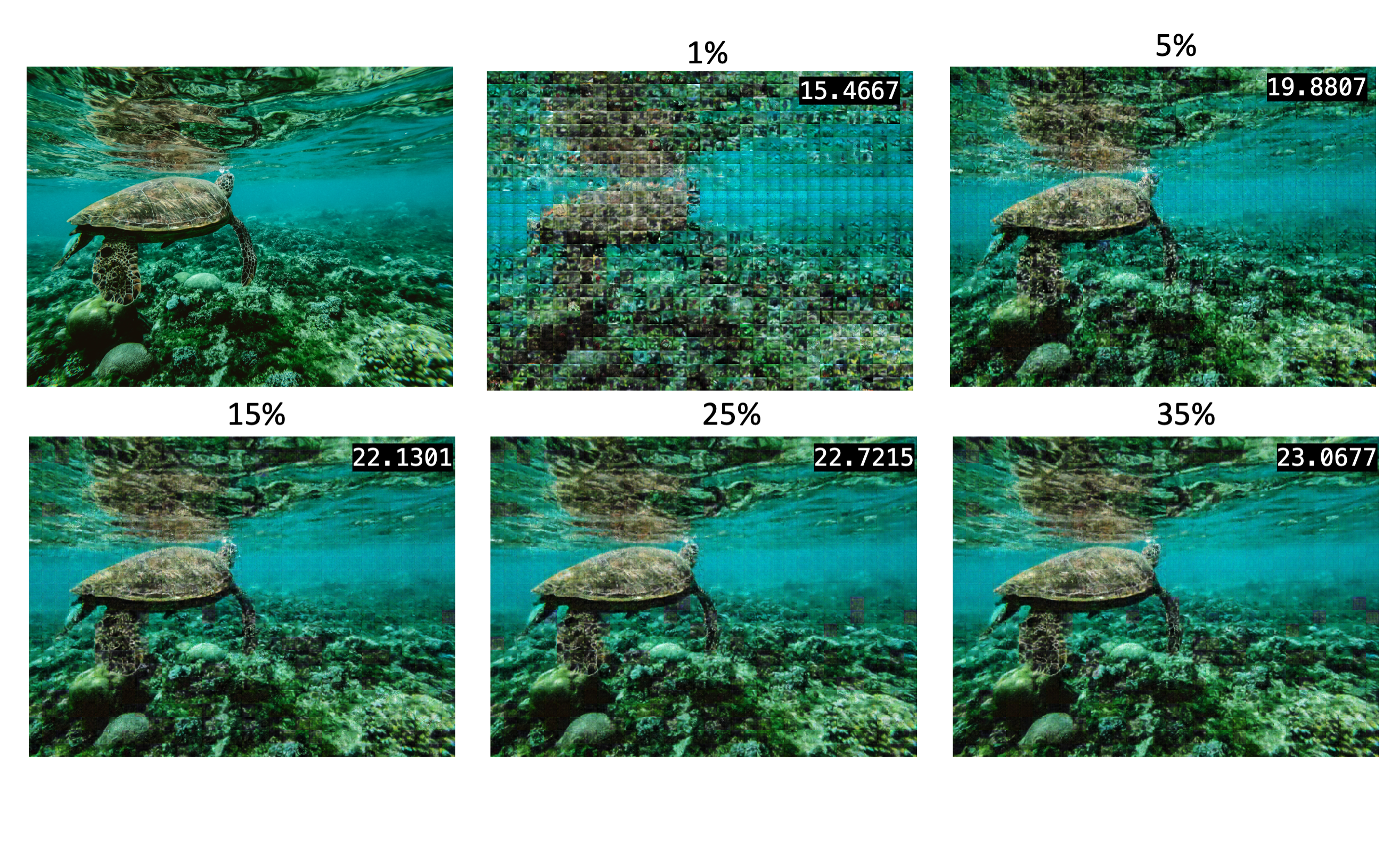}
   \caption{GPP degrades gracefully as the number of available observations are reduced. Results shown here are for an image of size $1024\times 768$, recovered using patches of size $32\times 32$. We do not use BM3D here to illustrate the patch artifacts under very few observations ($1\%$). The PSNR (dB) is also shown along with the reconstruction, compared to the ground truth which is shown in the top left.}
   \label{fig:cs_supp}
   \vspace{-10pt}
  \end{figure*}

\begin{figure*}[!htb]
  \centering
  \includegraphics[width=0.99\linewidth,clip,trim=0 0 0 0]{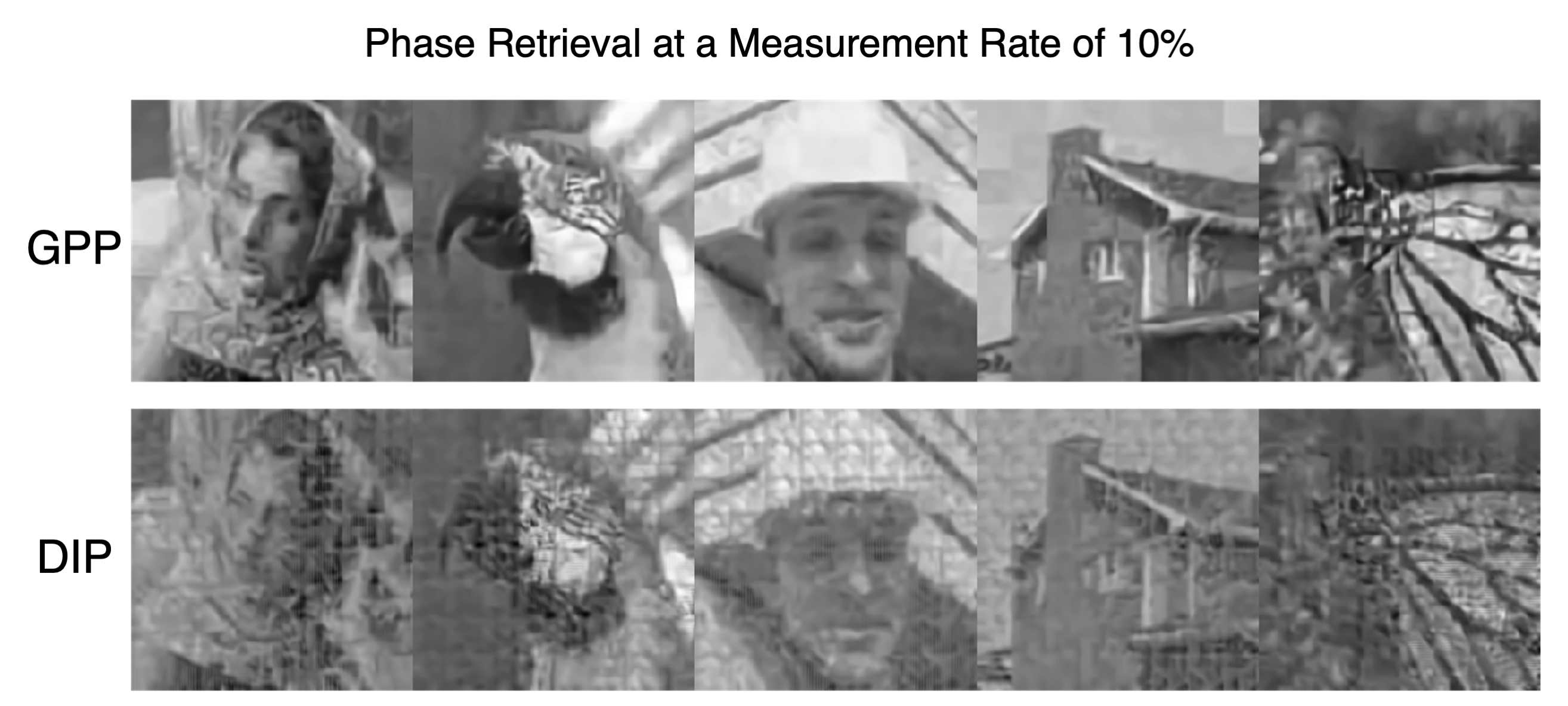}
   \caption{Compressive phase retrieval sensing at a measurement rate of $10\%$.}
   \label{fig:pr10}
   \vspace{-10pt}
  \end{figure*}

  \noindent In figure \ref{fig:pr10}, we show sample reconstructions for the phase retrieval task at a measurement rate of $10\%$.

  \subsection{Self calibration under unknown sensor shift}
\begin{figure*}[!htb]
  \centering
  \includegraphics[width=0.99\linewidth,clip,trim=0 0 0 0]{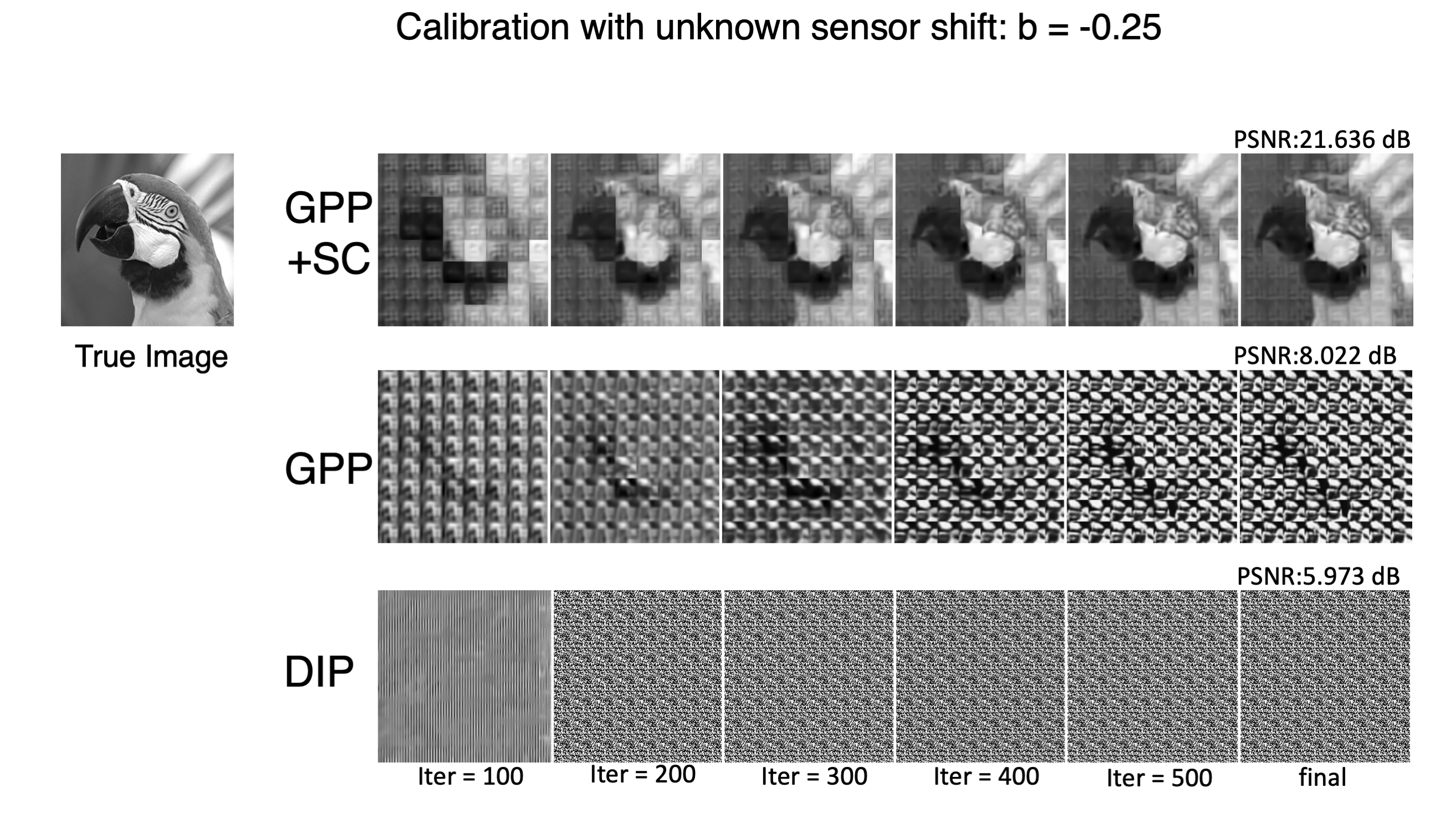}

   \caption{Compressive sensing at a measurement rate of $10\%$ under unknown sensor shift ($b$). We see that methods that do not account for this shift can easily break. Note the iterations in DIP are scaled since we run it for 10000 iterations compared to 1000 iterations on GPP and GPP + SC.}
   \label{fig:sc_shift}
   \vspace{-10pt}
  \end{figure*}

  In figure \ref{fig:sc_shift} we illustrate how reconstruction methods can easily fail to recover the solution when there is even a small shift in the operator. We simulate this using $b = -0.25$ and compare the proposed self calibration approach against no calibration and the untrained network prior (DIP) \cite{ulyanov2018deep}. We observe that the self calibration is able to successfully correct for the unknown shift, compared to the models that do not account for it.

  \section{Self-Calibrated Compressive Image Recovery}
\begin{figure*}[!htb]
      \centering
       \subfloat[unknown sensor gain (a)]{\includegraphics[width=0.3\linewidth]{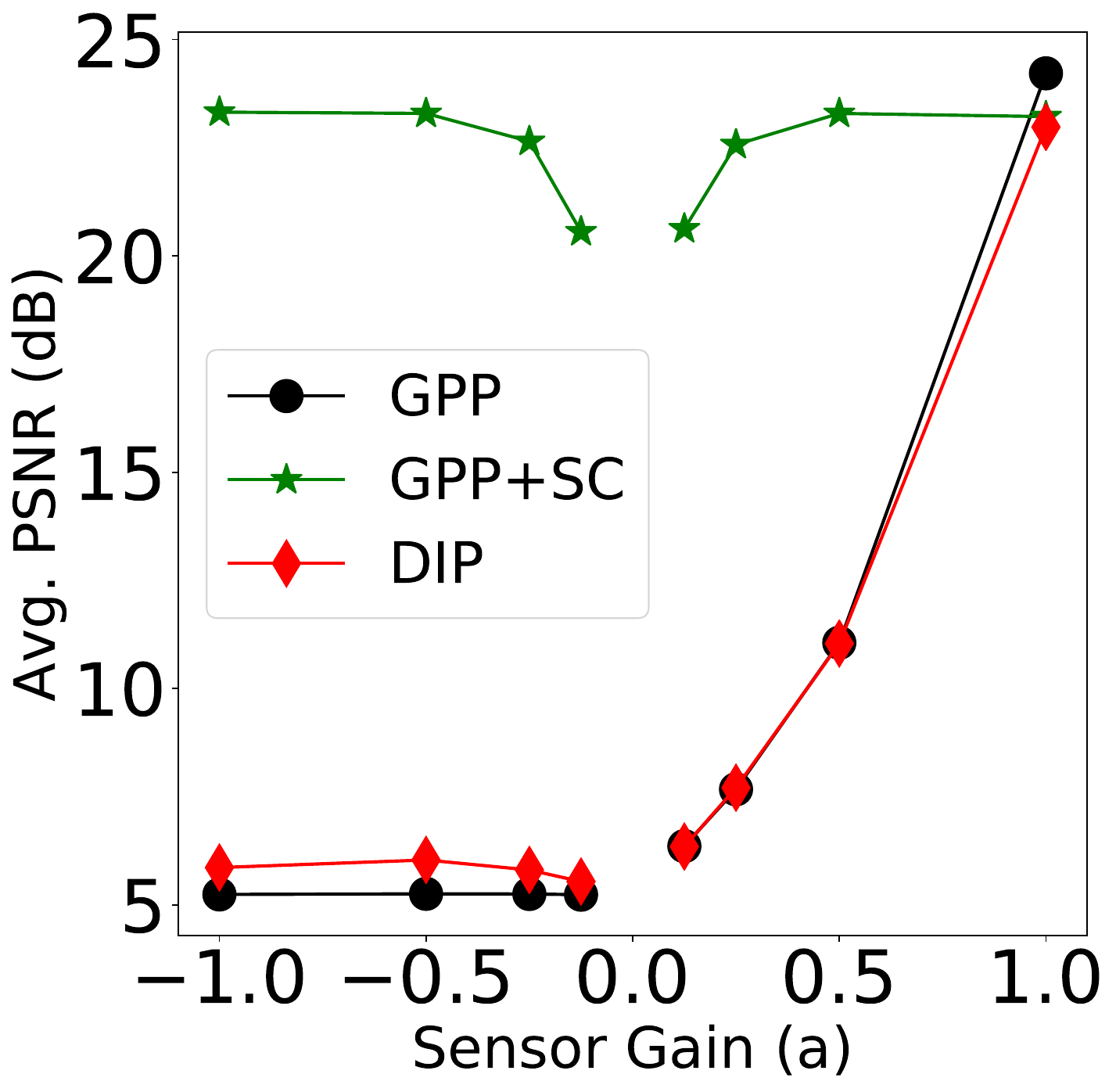}
       \label{fig:sensor_gain}}
        \subfloat[unknown sensor shift (b)]{\includegraphics[width=0.3\linewidth]{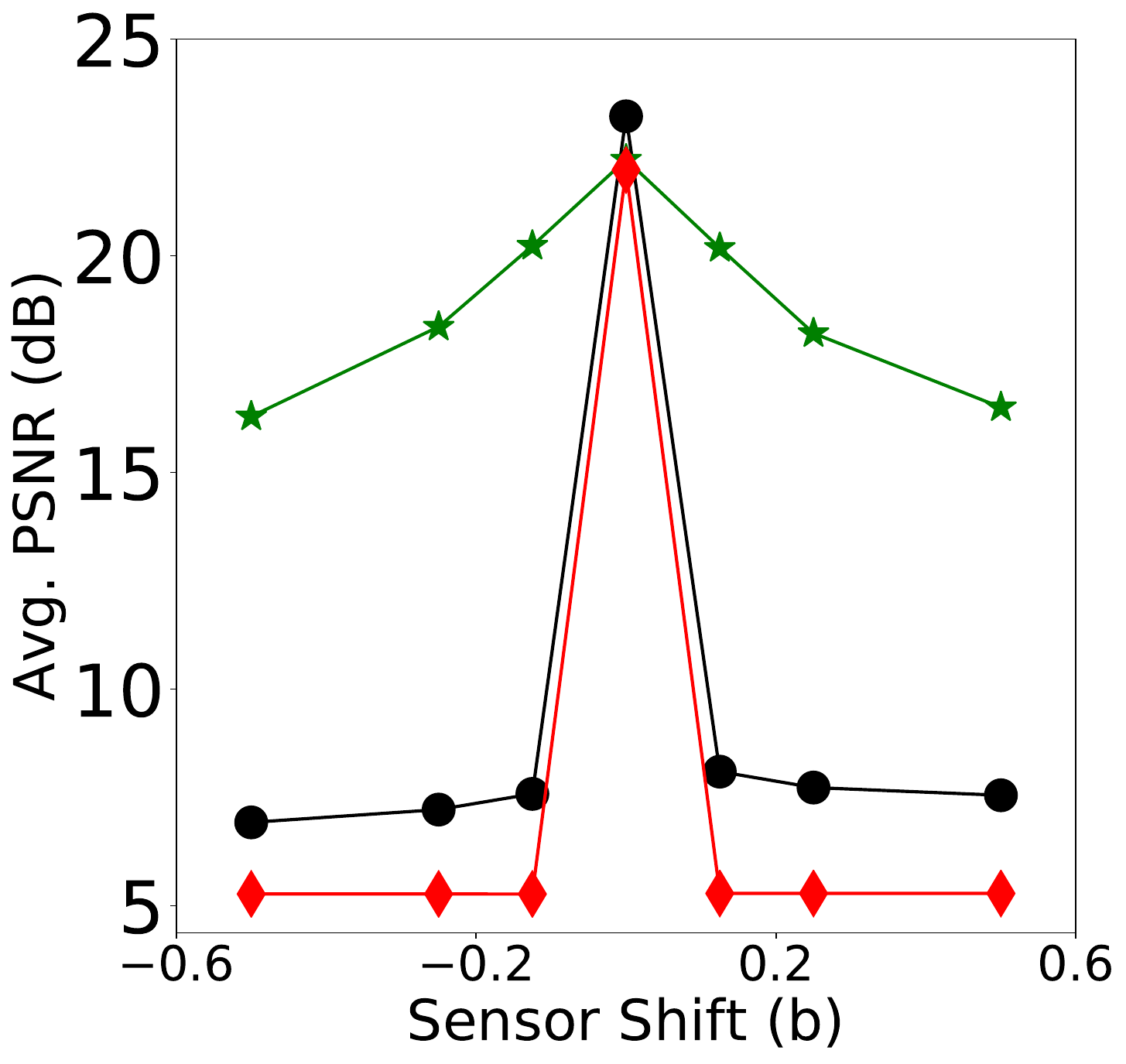}
       \label{fig:sensor_shift}}
       \subfloat[Convergence of
       the calibration algorithm]{\includegraphics[width=0.4\linewidth,clip,trim=10 10 10 0]{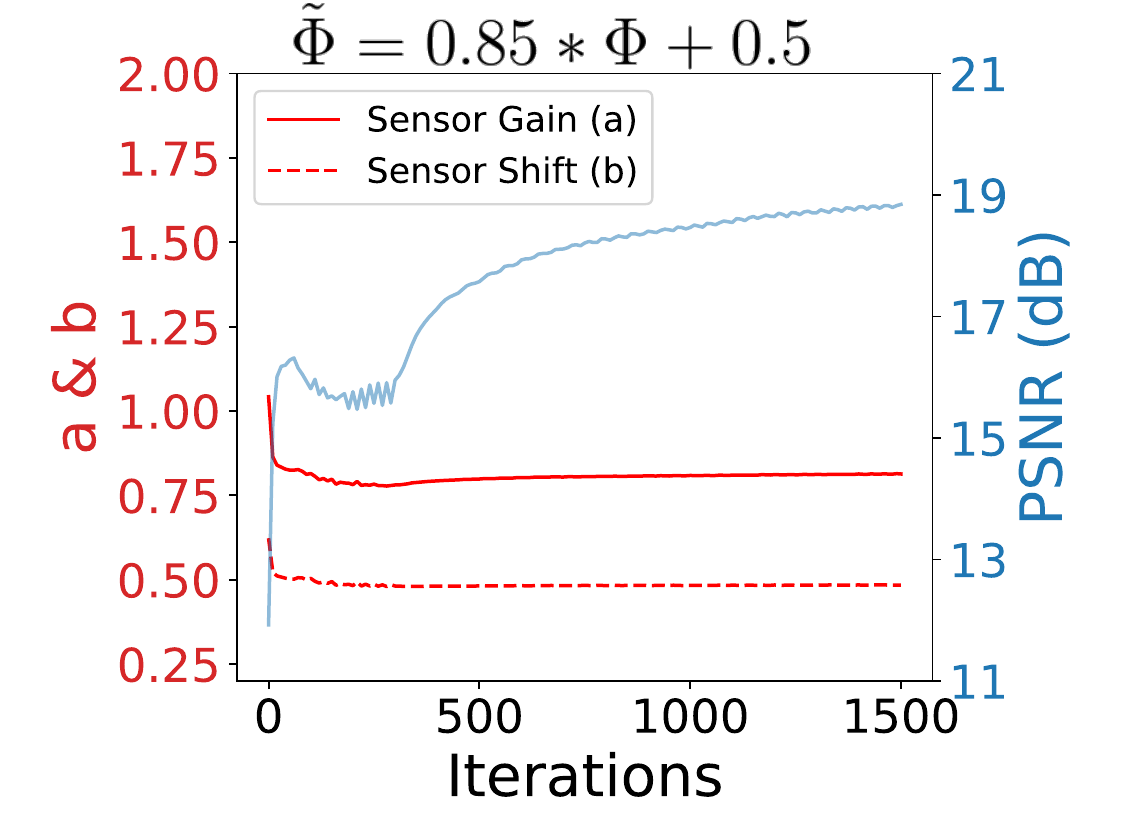}
       \label{fig:convergence}}
      \caption{Calibration experiments}
      \label{fig:calibration}
      \vspace{-10pt}
  \end{figure*}

  We evaluate the robustness of \name using the proposed self-calibration (SC) step. In this experiment, we perturb the measurement operator using the perturbation model described in section 4 of the main paper, $\tilde{\Phi}= a*\Phi+b$ using different values for $a$ and $b$. The measurements are then obtained by $\mathbf{y} = \tilde{\Phi}\mathbf{x}$, but all the reconstruction algorithms, including ours, are given access to only $\Phi$. We study the average PSNR for the seven test images used earlier, for different values of $a$, and $b$. In figure \ref{fig:sensor_gain}, we vary the gain coefficient $a$, while keeping $b$ fixed at 0.0. We observe that \name+SC remains robust to a wide variation of $a$, while the un-calibrated setup completely fails. We repeat these experiments for the sensor shift $b$ coefficient in figure \ref{fig:sensor_shift} where $a = 1.0$, and we vary $b$. We observe similarly that \name+SC is significantly more robust than \name, or DIP alone. Finally, in figure \ref{fig:convergence}, we study the convergence of the calibration algorithm for a mixed case with $a = 0.85, b = 0.5$. We see that the self-calibration step converges quickly to a value very close to the true values, and correspondingly improving the PSNR of the reconstruction.

  \section{Derivation for $a^*$ and $b^*$}
  Consider a vectorized square block of an image $\mathbf{x} \in \mathcal{X} \subset \mathbb{R}^n$ which we want to sense, and denote by $\mathbf{y} \in \mathbb{R}^m$ the compressive measurements obtained by the sensor. Given a measurement matrix $\Phi \in \mathbb{R}^{m \times n}$, with $m < n$ and $\Phi_{i,j} \sim \mathcal{N}(0,1)$, the compressive recovery problem is to estimate $\mathbf{x}$ accurately from $\mathbf{y}$. In the ideal setting, i.e., compressive sensing with known calibration the sensing model is given by $\mathbf{y} = \Phi \mathbf{x}$. Instead we consider a simple calibration model--- $\mathbf{y} = (a \Phi + b\mathbf{1})\mathbf{x}$, where $a,b \in \mathbb{R}^1$ are unknown calibration parameters and have to be estimated, and $\mathbf{1} \in \mathbb{R}^{m\times n}$ is a matrix of the same size as $\Phi$ with $1$s.

  In order to derive $a$ and $b$, we assume we have a current estimate of the solution $\x$ from a pre-trained generator $\mathcal{G}(\mathbf{z})$ for a latent vector $\mathbf{z}$. This is be randomly initialized at the beginning. Under this calibration model, let us define mean squared error loss function as follows:
  \begin{multline}
  \mathcal{L} = (\y - \pt\x)^\T(\y - \pt\x) \\
  \implies \mathcal{L} = \y^\T\y - \y^\T\pt\x - \x^\T\pt^\T\y + \x^\T\pt^\T\pt\x
  \end{multline}
  As a result, the derivatives with respect to each unknown $a,b$ are:

   \begin{equation}
   \begin{split}
   \label{eq:derivative_a_and_b}
   \frac{\partial \mathcal{L}}{\partial a} = -\y^\T\p\x - \x^T\p^\T\y + \x^\T\left[2a\p^\T\Phi+b\p^\T\1+b\1^\T\p \right]\x& \\
   \mbox{Similarly,~}\frac{\partial \mathcal{L}}{\partial b} =  - \y^\T\1\x - \x^\T\1^\T\y + \x^\T\left[a\p^\T\1 + \1^\T\p + 2b\1^\T\1\right]\x&
   \end{split}
   \end{equation}

   By setting these derivatives to zero, we get:
   \begin{align}
   \frac{\partial \mathcal{L}}{\partial a} = 0
   \implies -\y^\T\p\x -\x^\T\p^\T\y + 2a\x^\T\p^\T\p\x + b\x^\T\p^\T\1\x + b\x^\T\1^\T\p\x = 0.\label{b_0}\\
  \implies -2\y^\T\p\x + 2a\x^\T\p^\T\p\x + 2b\x^\T\p^\T\1\x = 0.\label{b_1}\\
  \implies  b = \frac{\y^\T\p\x - a\x^\T\p^\T\p\x}{\x^\T\p^\T\1\x}\label{b_2}
  \end{align}
  Note, in \eqref{b_1} all the terms are scalars and therefore $\y^\T\p\x =\x^\T\p^\T\y$ etc. Next, we take the partial derivative with respect to $b$.
  \begin{align}
   \frac{\partial \mathcal{L}}{\partial b} = 0
   \implies \y^\T\1\x -\x^\T\1^\T\y + a\x^\T\p^\T\1\x+a\x^\T\1^\T\p\x + 2b\x^\T\1^\T\1\x = 0. \label{a_0} \\
   \implies -2\y^\T\1\x + 2a\x^\T\p^\T\1\x + 2b\x^\T\1^\T\1\x = 0. \label{a_1}\\
   \implies b = \frac{\y^T\1\x - a\x^\T\p^\T\1\x}{\x^\T\1^\T\1\x} \label{a_2}
  \end{align}
  Combining equations \eqref{a_2} and \eqref{b_2}, we get the following:
  \begin{align}
  \label{ab_0}
  \left(\y^\T\p\x - a\x^\T\p^\T\p\x\right)\x^\T\1^\T\1\x = \left(\y^T\1\x - a\x^\T\p^\T\1\x\right)\x^\T\p^\T\1\x
  \end{align}
  As in the paper, let us define scalar quantities for notational convenience: $c_{\Phi} = \mathbf{y}^T\Phi\mathbf{x}, c_{\mathbf{1}} = \mathbf{y}^T\mathbf{1}\mathbf{x},  \theta_{\Phi} = (\Phi\mathbf{x})^T(\Phi\mathbf{x}), \theta_{\mathbf{1}} = (\mathbf{1}\mathbf{x})^T(\mathbf{1}\mathbf{x}), \lambda = (\Phi\mathbf{x})^T(\mathbf{1}\mathbf{x})$. This implies, \eqref{ab_0} is now reformulated as:
  \begin{align}
  (c_\Phi - a\theta_\Phi)\theta_\mathbf{1} = (c_\mathbf{1}-a\lambda)\lambda\\
  \implies a = \frac{c_\mathbf{1}\lambda-c_{\Phi}\theta_\mathbf{1}}{\lambda^2-\theta_\Phi\theta_\mathbf{1}}\label{a_star}\\
  \mbox{~and~} b = \frac{c_{\mathbf{1}}-a^*\lambda}{\theta_{\mathbf{1}}}\label{b_star}
  \end{align}
  In each step of the alternating minimization, we use the estimates from \eqref{a_star}, and \eqref{b_star} and update the latent vector $\mathbf{z}$, which is repeated until convergence in $a,b,\mathbf{z}$. Since our generative model is defined at a patch level, we estimate $a,b$ for each individual patch $i$ separately and assign the mean values of all the patch-estimates as the single $a,b$ for the entire image: $a = \frac{1}{N}\sum_{i=1}^N a_i$; and $b = \frac{1}{N}\sum_{i=1}^N b_i$. We continue with this alternating minimization until the loss converges. In practice, we find that the algorithm converges within 1500 iterations, and finding recovery and convergence properties of this algorithm remain part of our future work. We empirically study convergence properties of this self-calibrating mechanism under different settings in the supplement.
  \begin{figure*}[!htb]
  \centering
  \includegraphics[width=0.99\linewidth,clip,trim=0 0 0 0]{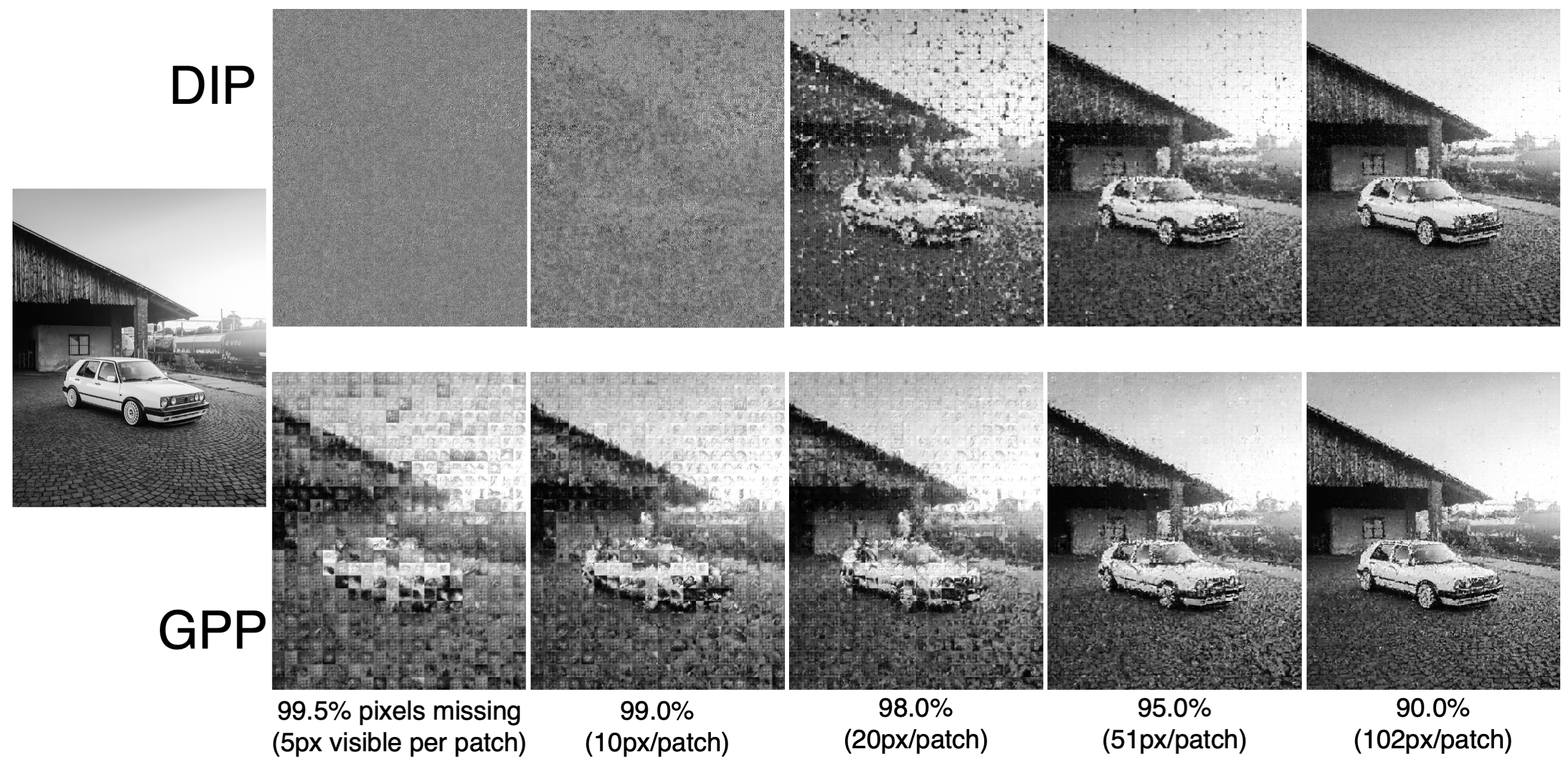}
   \caption{\textbf{GPP for image inpainting:} GPP is more efficient than DIP, being able to recover parts of original image even when 99.5
   \% of the pixels are missing. The original image is of size $800\times640$.}
   \label{fig:inpainting}
   \vspace{-10pt}
  \end{figure*}
  \paragraph{}Note that the latent space optimization is itself a nonconvex optimization problem which is solved only approximately using a gradient-descent type of optimization, yielding a local minimum at each iteration. Therefore the overall optimization problem is nonconvex even though the calibration parameters can be estimated exactly at each step. Using the result in Theorem 3.1 and Lemma 3.2 in the paper by beck \cite{beck2015convergence}, we can see that the alternating procedure converges to a stationary point. A stronger result is deferred for future work.

  \subsection{GPP for image inpainting}
  GPP is a generic prior to constrain solutions to the natural image manifold. We show an example here of how it can be used in other challenging inverse problems. In figure \ref{fig:inpainting}, we illustrate the efficiency of GPP for a for inpainting, where only a small number random pixels are shown, and the task is to recover the original image. Unlike most existing methods, we see that GPP's solution degrades more gracefully than DIP, even recovering some signal when 99.5\% of the pixels are missing.
\end{appendices}

\end{document}


\title{\emph{Supplementary Material}: Generative Patch Priors for Practical Compressive Image Recovery}
\maketitle

\begin{figure*}[!htb]
\centering
\includegraphics[width=0.99\linewidth,clip,trim=0 30 0 0]{figs/supp_figure.png}
 \caption{GPP degrades gracefully as the number of available observations are reduced. Results shown here are for an image of size $1024\times 768$, recovered using patches of size $32\times 32$. We do not use BM3D here to illustrate the patch artifacts under very few observations ($1\%$). The PSNR (dB) is also shown along with the reconstruction, compared to the ground truth which is shown in the top left.}
 \label{fig:cs_supp}
 \vspace{-10pt}
\end{figure*}
\section{Additional results}
\noindent In figure \ref{fig:pr10}, we show sample reconstructions for the phase retrieval task at a measurement rate of $10\%$.
\begin{figure*}[!htb]
\centering
\includegraphics[width=0.99\linewidth,clip,trim=0 0 0 0]{figs/pr_results_10.png}
 \caption{Compressive phase retrieval sensing at a measurement rate of $10\%$.}
 \label{fig:pr10}
 \vspace{-10pt}
\end{figure*}

\subsection{Self calibration under unknown sensor shift}
In figure \ref{fig:sc_shift} we illustrate how reconstruction methods can easily fail to recover the solution when there is even a small shift in the operator. We simulate this using $b = -0.25$ and compare the proposed self calibration approach against no calibration and the untrained network prior (DIP) \cite{ulyanov2018deep}. We observe that the self calibration is able to successfully correct for the unknown shift, compared to the models that do not account for it.
\begin{figure*}[!htb]
\centering
\includegraphics[width=0.99\linewidth,clip,trim=0 0 0 0]{figs/sample_unknown_shift.png}

 \caption{Compressive sensing at a measurement rate of $10\%$ under unknown sensor shift ($b$). We see that methods that do not account for this shift can easily break. Note the iterations in DIP are scaled since we run it for 10000 iterations compared to 1000 iterations on GPP and GPP + SC.}
 \label{fig:sc_shift}
 \vspace{-10pt}
\end{figure*}

\section{Self-Calibrated Compressive Image Recovery}
We evaluate the robustness of \name using the proposed self-calibration (SC) step. In this experiment, we perturb the measurement operator using the perturbation model described in section 4 of the main paper, $\tilde{\Phi}= a*\Phi+b$ using different values for $a$ and $b$. The measurements are then obtained by $\mathbf{y} = \tilde{\Phi}\mathbf{x}$, but all the reconstruction algorithms, including ours, are given access to only $\Phi$. We study the average PSNR for the seven test images used earlier, for different values of $a$, and $b$. In figure \ref{fig:sensor_gain}, we vary the gain coefficient $a$, while keeping $b$ fixed at 0.0. We observe that \name+SC remains robust to a wide variation of $a$, while the un-calibrated setup completely fails. We repeat these experiments for the sensor shift $b$ coefficient in figure \ref{fig:sensor_shift} where $a = 1.0$, and we vary $b$. We observe similarly that \name+SC is significantly more robust than \name, or DIP alone. Finally, in figure \ref{fig:convergence}, we study the convergence of the calibration algorithm for a mixed case with $a = 0.85, b = 0.5$. We see that the self-calibration step converges quickly to a value very close to the true values, and correspondingly improving the PSNR of the reconstruction.

\begin{figure*}[!htb]
    \centering
     \subfloat[unknown sensor gain (a)]{\includegraphics[width=0.3\linewidth]{figs/calibration_a.pdf}
     \label{fig:sensor_gain}}
      \subfloat[unknown sensor shift (b)]{\includegraphics[width=0.3\linewidth]{figs/calibration_b.pdf}
     \label{fig:sensor_shift}}
     \subfloat[Convergence of
     the calibration algorithm]{\includegraphics[width=0.4\linewidth,clip,trim=10 10 10 0]{figs/convergence.pdf}
     \label{fig:convergence}}
    \caption{Calibration experiments}
    \label{fig:calibration}
    \vspace{-10pt}
\end{figure*}

\section{Derivation for $a^*$ and $b^*$}
Consider a vectorized square block of an image $\mathbf{x} \in \mathcal{X} \subset \mathbb{R}^n$ which we want to sense, and denote by $\mathbf{y} \in \mathbb{R}^m$ the compressive measurements obtained by the sensor. Given a measurement matrix $\Phi \in \mathbb{R}^{m \times n}$, with $m < n$ and $\Phi_{i,j} \sim \mathcal{N}(0,1)$, the compressive recovery problem is to estimate $\mathbf{x}$ accurately from $\mathbf{y}$. In the ideal setting, i.e., compressive sensing with known calibration the sensing model is given by $\mathbf{y} = \Phi \mathbf{x}$. Instead we consider a simple calibration model--- $\mathbf{y} = (a \Phi + b\mathbf{1})\mathbf{x}$, where $a,b \in \mathbb{R}^1$ are unknown calibration parameters and have to be estimated, and $\mathbf{1} \in \mathbb{R}^{m\times n}$ is a matrix of the same size as $\Phi$ with $1$s.

In order to derive $a$ and $b$, we assume we have a current estimate of the solution $\x$ from a pre-trained generator $\mathcal{G}(\mathbf{z})$ for a latent vector $\mathbf{z}$. This is be randomly initialized at the beginning. Under this calibration model, let us define mean squared error loss function as follows:
\begin{multline}
\mathcal{L} = (\y - \pt\x)^\T(\y - \pt\x) \\
\implies \mathcal{L} = \y^\T\y - \y^\T\pt\x - \x^\T\pt^\T\y + \x^\T\pt^\T\pt\x
\end{multline}
As a result, the derivatives with respect to each unknown $a,b$ are:

 \begin{equation}
 \begin{split}
 \label{eq:derivative_a_and_b}
 \frac{\partial \mathcal{L}}{\partial a} = -\y^\T\p\x - \x^T\p^\T\y + \x^\T\left[2a\p^\T\Phi+b\p^\T\1+b\1^\T\p \right]\x& \\
 \mbox{Similarly,~}\frac{\partial \mathcal{L}}{\partial b} =  - \y^\T\1\x - \x^\T\1^\T\y + \x^\T\left[a\p^\T\1 + \1^\T\p + 2b\1^\T\1\right]\x&
 \end{split}
 \end{equation}

 By setting these derivatives to zero, we get:
 \begin{align}
 \frac{\partial \mathcal{L}}{\partial a} = 0
 \implies -\y^\T\p\x -\x^\T\p^\T\y + 2a\x^\T\p^\T\p\x + b\x^\T\p^\T\1\x + b\x^\T\1^\T\p\x = 0.\label{b_0}\\
\implies -2\y^\T\p\x + 2a\x^\T\p^\T\p\x + 2b\x^\T\p^\T\1\x = 0.\label{b_1}\\
\implies  b = \frac{\y^\T\p\x - a\x^\T\p^\T\p\x}{\x^\T\p^\T\1\x}\label{b_2}
\end{align}
Note, in \eqref{b_1} all the terms are scalars and therefore $\y^\T\p\x =\x^\T\p^\T\y$ etc. Next, we take the partial derivative with respect to $b$.
\begin{align}
 \frac{\partial \mathcal{L}}{\partial b} = 0
 \implies \y^\T\1\x -\x^\T\1^\T\y + a\x^\T\p^\T\1\x+a\x^\T\1^\T\p\x + 2b\x^\T\1^\T\1\x = 0. \label{a_0} \\
 \implies -2\y^\T\1\x + 2a\x^\T\p^\T\1\x + 2b\x^\T\1^\T\1\x = 0. \label{a_1}\\
 \implies b = \frac{\y^T\1\x - a\x^\T\p^\T\1\x}{\x^\T\1^\T\1\x} \label{a_2}
\end{align}
Combining equations \eqref{a_2} and \eqref{b_2}, we get the following:
\begin{align}
\label{ab_0}
\left(\y^\T\p\x - a\x^\T\p^\T\p\x\right)\x^\T\1^\T\1\x = \left(\y^T\1\x - a\x^\T\p^\T\1\x\right)\x^\T\p^\T\1\x
\end{align}
As in the paper, let us define scalar quantities for notational convenience: $c_{\Phi} = \mathbf{y}^T\Phi\mathbf{x}, c_{\mathbf{1}} = \mathbf{y}^T\mathbf{1}\mathbf{x},  \theta_{\Phi} = (\Phi\mathbf{x})^T(\Phi\mathbf{x}), \theta_{\mathbf{1}} = (\mathbf{1}\mathbf{x})^T(\mathbf{1}\mathbf{x}), \lambda = (\Phi\mathbf{x})^T(\mathbf{1}\mathbf{x})$. This implies, \eqref{ab_0} is now reformulated as:
\begin{align}
(c_\Phi - a\theta_\Phi)\theta_\mathbf{1} = (c_\mathbf{1}-a\lambda)\lambda\\
\implies a = \frac{c_\mathbf{1}\lambda-c_{\Phi}\theta_\mathbf{1}}{\lambda^2-\theta_\Phi\theta_\mathbf{1}}\label{a_star}\\
\mbox{~and~} b = \frac{c_{\mathbf{1}}-a^*\lambda}{\theta_{\mathbf{1}}}\label{b_star}
\end{align}
In each step of the alternating minimization, we use the estimates from \eqref{a_star}, and \eqref{b_star} and update the latent vector $\mathbf{z}$, which is repeated until convergence in $a,b,\mathbf{z}$. Since our generative model is defined at a patch level, we estimate $a,b$ for each individual patch $i$ separately and assign the mean values of all the patch-estimates as the single $a,b$ for the entire image: $a = \frac{1}{N}\sum_{i=1}^N a_i$; and $b = \frac{1}{N}\sum_{i=1}^N b_i$. We continue with this alternating minimization until the loss converges. In practice, we find that the algorithm converges within 1500 iterations, and finding recovery and convergence properties of this algorithm remain part of our future work. We empirically study convergence properties of this self-calibrating mechanism under different settings in the supplement. 
\begin{figure*}[!htb]
\centering
\includegraphics[width=0.99\linewidth,clip,trim=0 0 0 0]{figs/inpainting.png}
 \caption{\textbf{GPP for image inpainting:} GPP is more efficient than DIP, being able to recover parts of original image even when 99.5
 \% of the pixels are missing. The original image is of size $800\times640$.}
 \label{fig:inpainting}
 \vspace{-10pt}
\end{figure*}
\paragraph{}Note that the latent space optimization is itself a nonconvex optimization problem which is solved only approximately using a gradient-descent type of optimization, yielding a local minimum at each iteration. Therefore the overall optimization problem is nonconvex even though the calibration parameters can be estimated exactly at each step. Using the result in Theorem 3.1 and Lemma 3.2 in the paper by beck \cite{beck2015convergence}, we can see that the alternating procedure converges to a stationary point. A stronger result is deferred for future work.

\subsection{GPP for image inpainting}
GPP is a generic prior to constrain solutions to the natural image manifold. We show an example here of how it can be used in other challenging inverse problems. In figure \ref{fig:inpainting}, we illustrate the efficiency of GPP for a for inpainting, where only a small number random pixels are shown, and the task is to recover the original image. Unlike most existing methods, we see that GPP's solution degrades more gracefully than DIP, even recovering some signal when 99.5\% of the pixels are missing.

\small{
\bibliographystyle{ieee_fullname}
\bibliography{references}}